\documentclass[lettersize,journal]{IEEEtran}
\usepackage{amsmath,amsfonts}
\usepackage{algorithmic}
\usepackage{algorithm}
\usepackage{array}
\usepackage{textcomp}
\usepackage{stfloats}
\usepackage{url}
\usepackage{verbatim}
\usepackage{graphicx}
\usepackage{cleveref}
\usepackage{multirow}
\usepackage{tcolorbox}
\usepackage{titlesec}
\usepackage{diagbox} 
\usepackage{makecell} 
\usepackage{letltxmacro}
\usepackage[dvipsnames]{xcolor}
\usepackage[table,xcdraw]{xcolor}
\usepackage{colortbl}
\usepackage{pifont}
\usepackage[edges]{forest} 
\usepackage{xcolor}        
\usepackage{booktabs}

\begin{document}

\title{LLM-based Agents Suffer from Hallucinations: A Survey of Taxonomy, Methods, and Directions}

\author{Xixun Lin,
        Yucheng Ning,
        Jingwen Zhang,
        Yan Dong,
        Yilong Liu,
        Yongxuan Wu,
        Xiaohua Qi,
        Nan Sun,
        Yanmin Shang,
        Kun Wang,
        Pengfei Cao,
        Qingyue Wang, 
        Lixin Zou,
        Xu Chen,
        Chuan Zhou,
        Jia Wu,
        Peng Zhang,
        Qingsong Wen, 
        Shirui Pan,
        Bin Wang,
        Yanan Cao,
        Kai Chen,
        Songlin Hu,
        Li Guo\
\IEEEcompsocitemizethanks{\IEEEcompsocthanksitem X. Lin, Y. Ning, J. Zhang, Y. Dong, Y. Liu, Y. Wu, X. Qi, N. Sun, Y. Shang, Y. Cao, K. Chen, S. Hu, and L. Guo are with Institute of Information Engineering, Chinese Academy of Sciences, School of Cyber Security, University of Chinese Academy of Sciences, Beijing, China. 
\IEEEcompsocthanksitem K. Wang is with Nanyang Technological University, Singapore. 
\IEEEcompsocthanksitem P. Cao is with Institute of Automation, Chinese Academy of Sciences, Beijing, China. 
\IEEEcompsocthanksitem Q. Wang is with Hong Kong University of Science and Technology, Hong Kong, China. 
\IEEEcompsocthanksitem L. Zou is with School of Cyber Science and Engineering, Wuhan University, Wuhan, China. 
\IEEEcompsocthanksitem X. Chen is with Gaoling School of Artificial Intelligence, Renmin University of China, Beijing, China. 
\IEEEcompsocthanksitem C. Zhou is with Academy of Mathematics and Systems Science, Chinese Academy of Sciences, Beijing, China.
\IEEEcompsocthanksitem J. Wu is with School of Computing, Faculty of Science and Engineering, Macquarie University, Sydney, Australia. 
\IEEEcompsocthanksitem P. Zhang is with the Cyberspace Institute of Advanced Technology, Guangzhou University, Guangzhou, China. 
\IEEEcompsocthanksitem Q. Wen is with Squirrel Ai Learning, Bellevue, USA.
\IEEEcompsocthanksitem S. Pan is with School of Information and Communication Technology, Griffith University, Gold Coast, Australia.
\IEEEcompsocthanksitem B. Wang is with Xiaomi Company, Beijing, China. 
}
}

\markboth{Journal of \LaTeX\ Class Files,~Vol.~14, No.~8, August~2021}%
{Shell \MakeLowercase{\textit{et al.}}: A Sample Article Using IEEEtran.cls for IEEE Journals}

\IEEEpubid{0000--0000/00\$00.00~\copyright~2021 IEEE}

\maketitle

\begin{abstract}
Driven by the rapid advancements of Large Language Models (LLMs), LLM-based agents have emerged as powerful intelligent systems capable of human-like cognition, reasoning, and interaction. These agents are increasingly being deployed across diverse real-world applications, including student education, scientific research, and financial analysis. However, despite their remarkable potential, LLM-based agents remain vulnerable to hallucination issues, which can result in erroneous task execution and undermine the reliability of the overall system design. Addressing this critical challenge requires a deep understanding and a systematic consolidation of recent advances on LLM-based agents. To this end, we present the first comprehensive survey of hallucinations in LLM-based agents. By carefully analyzing the complete workflow of agents, we propose a new taxonomy that identifies different types of agent hallucinations occurring at different stages. Furthermore, we conduct an in-depth examination of eighteen triggering causes underlying the emergence of agent hallucinations. Through a detailed review of a large number of existing studies, we summarize approaches for hallucination mitigation and detection, and highlight promising directions for future research. We hope this survey will inspire further efforts toward addressing hallucinations in LLM-based agents, ultimately contributing to the development of more robust and reliable agent systems.
\end{abstract}


\section{Introduction}

\IEEEPARstart{L}Language Models (LLMs)~\cite{achiam2023gpt,touvron2023llama,touvron2023llama2,grattafiori2024llama,guo2025deepseek} have recently showcased extraordinary capabilities across a broad spectrum of tasks, including language generation~\cite{wang2023grammar}, intent comprehension~\cite{geng2024large}, and knowledge reasoning~\cite{luo2023reasoning}. These capabilities are largely attributed to the vast scale of training data~\cite{yu2023large}, model architecture innovations~\cite{vaswani2017attention}, and emergent abilities~\cite{wei2022emergent} that arise during instruction tuning~\cite{longpre2023flan} and in-context learning~\cite{dong2022survey}. Building on these breakthroughs, \textbf{LLM-based Agents}~\cite{yao2023react,shinn2023reflexion,chen2024agent,wu2024graph,wang2024survey} are becoming increasingly proficient in task automation across a wide range of fields, marking a critical milestone on the path toward Artificial General Intelligence (AGI)~\cite{goertzel2014artificial}. Moreover, these agents can be orchestrated into a \textbf{LLM-based Multi-agent System} (MAS)~\cite{park2023generative,li2023camel,hong2024metagpt,chan2023chateval,guo2024large}, where different agents with distinct specializations collaborate and interact to solve real-world and complex problems beyond the capacity of any single agent through mutual cooperation, such as knowledge sharing~\cite{ba2024cautiously} and collaborative coordination~\cite{zhu2025multiagentbench}.
\par
Despite the impressive performance achieved by LLM-based agents, their rapid advancement introduces a spectrum of safety challenges~\cite{zhang2024agent,yuan2024r,tian2023evil}. Among these challenges, \textbf{Agent Hallucinations} represents a particularly significant threat~\cite{liu2025advances,r55,deng2025ai}. Previous studies on hallucinations have primarily focused on the field of natural language generation (NLG)~\cite{dong2022survey}, where hallucinations refer to the phenomenon in which NLG models generate unfaithful or nonsensical text. 
Ji et al. review recent advances in addressing hallucinations across various NLG tasks~\cite{ji2023survey}. Meanwhile, Huang et al. specifically focus on the causes of LLM hallucinations~\cite{huang2025survey}. In this context, hallucinations in LLMs are categorized into factuality and faithfulness hallucinations. Factuality hallucinations highlight discrepancies between the generated content and verifiable real-world facts~\cite{min2023factscore}; while faithfulness hallucinations refer to deviations from the user’s original input~\cite{fabbri2021qafacteval}. In addition, the authors discuss several effective methods for detecting and mitigating hallucination issues~\cite{liu2025reducing}.
\par 
Unlike the aforementioned settings, the LLM-based agent is a more sophisticated intelligent system, equipped with goal-directed reasoning and action-taking capabilities. Such agents typically comprise three core modules: brain, perception, and action~\cite{xi2025rise}.
\IEEEpubidadjcol
The brain module is primarily responsible for storing memory and knowledge, supporting reasoning and decision-making for executing tasks; the perception module extends the agent’s perceptual space, enabling it to handle multi-modal environmental inputs; and the action module expands the agent’s action space, allowing it not only to generate textual outputs but also to invoke tools for completing more complex tasks. Therefore, in LLM-based agents, hallucinations are not "linguistic errors", but rather a broad category of fabricated or misjudged "human-like behaviors" that may occur at any stage of the agent’s pipeline. Accordingly, the manifestations and causes of agent hallucinations are considerably more complex. This complexity is reflected in three key aspects: \textbf{1) More Diverse Types}: Rather than the straightforward response errors of a single model, agent hallucinations are compound behaviors arising from interactions among multiple modules, resulting in a broader and more varied range of hallucination types. \textbf{2) Longer Propagation Chains}: The aforementioned hallucinations are mostly localized and single-step errors, whereas agent hallucinations often span multiple steps and involve multi-state transitions. Such hallucinations are not limited to the final output; they may also arise during intermediate processes such as perception and reasoning, where they can propagate and accumulate over time. \textbf{3) More Severe Consequences}: Agent hallucinations involve "physically consequential" errors, where incorrect embodied actions can directly affect task execution, system devices, and user experiences in the real world. As a result, the cost and risk associated with agent hallucinations are significantly higher. However, existing reviews of LLM-based agents primarily focus on architectural designs and practical applications, while giving far insufficient attention to the importance and urgency of agent hallucinations.

\begin{figure*}[htbp]
  \centering
  \includegraphics[width=1.0\textwidth]{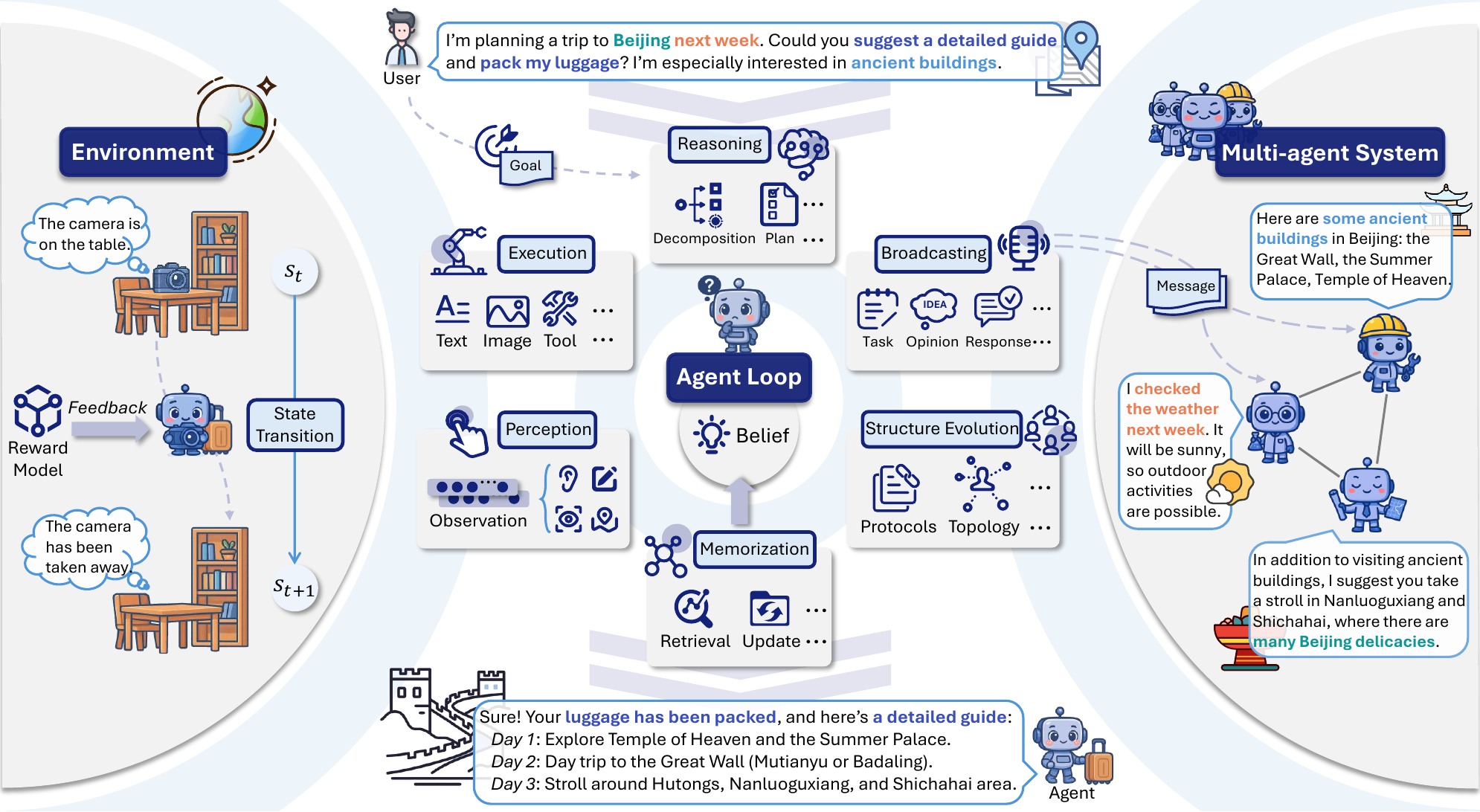}
  \caption{An overview of agent goal completion. Within the loop, the LLM-based agent carries out external behaviors such as reasoning, execution, perception, and memorization, guided by its internal belief state. Throughout this process, the environment dynamically evolves in response to the agent’s decisions, while task allocation within the LLM-based multi-agent system including broadcasting and structure evolution further enhances the fulfillment of user requirements.}
  \label{fig:agent_loop}
\end{figure*}

\par
To this end, we provide an overview of agent hallucinations to fill this important gap and promote the advancement of agents. In this paper,  the interaction dynamics of LLM-based agents are formulated as a partially observable Markov decision process (POMDP) where the agent\footnote{Throughout this paper, we use the term "agent" to denote the LLM-based agent, unless otherwise specified.} interacts with the learning environment, makes decisions, receives feedback, and updates its state over multiple time steps. Based on this general POMDP setting, we summarize the contributions of our work below. 

\begin{itemize}

\item \textbf{First Survey.} To the best of our knowledge, this is the first survey to review hallucination issues in LLM-based agents. It encompasses recent research on both mitigation and detection methods, offering a broad perspective on the development of LLM-based agents.

\item \textbf{Innovative Taxonomy.} We introduce a novel decomposition of agent components into two parts: internal state and external behaviors. The former is characterized by a belief state maintained by the agent, which serves as the most fundamental unit of the agent cognition policy. External behaviors refer to a series of proactive procedures guided by the belief state. Through this internal–external distinction, we can classify agent hallucinations based on the specific stage at which they occur, covering five types of agent hallucinations.

\item \textbf{Comprehensive Review.} For each type of agent hallucinations, we offer a formal definition, illustrative examples, and an in-depth discussion of representative studies. Based on this, we identify eighteen triggering causes of agent hallucinations. Furthermore, we summarize ten general approaches for hallucination mitigation, along with corresponding detection methods, to provide readers  with a clear and up-to-date overview of the current research landscape on agent hallucinations.  

\item \textbf{Future Outlook.} By reviewing and summarizing existing solutions to hallucination problems and their possible limitations, we outline several promising directions for future exploration that will need to be fully investigated to advance both academic research and real-world deployment of LLM-based agents.

\item \textbf{Open Resource.} We provide a well-curated collection of resources, encompassing over 200+ related papers, which we make publicly available on GitHub\footnote{https://github.com/ASCII-LAB/Awesome-Agent-Hallucinations.} to foster community engagement and collaboration.
\end{itemize}

\textbf{Paper Organization.}
The remainder of this paper is structured as follows. Section~\ref{section_2} introduces the formal definition of LLM-based agents. Based on this definition,  Section~\ref{section_3} presents a new taxonomy of agent hallucinations. Section~\ref{Systematic_Methodology} reviews existing solutions, with particular emphasis on methodologies for hallucination mitigation. Section~\ref{section_5} outlines future research directions. Section~\ref{section_6} concludes the survey.

\section{Formal Definition of LLM-based Agents}
\label{section_2}
Fig.~\ref{fig:agent_loop} provides an overview of agent goal completion. The formal definition of LLM-based agents including interaction dynamics and LLM-based agent loop is given here.

\subsection{Interaction Dynamics}
The interaction dynamics between the agent and the learning environment is usually formulated as a Partially Observable Markov Decision Process (POMDP)~\cite{aastrom1965optimal} which is defined as a 8-tuple: $\mathcal{E} = (\mathcal{S},\mathcal{A},T,\mathcal{G}, \mathcal{O},Z,R,\gamma)$. The description of each element in $\mathcal{E}$ is given here: 

\begin{itemize}

\item $\mathcal{S}$ is the state space where $s \in \mathcal{S}$ represents a true environment state. A POMDP explicitly models an agent decision process in which the agent cannot directly observe the underlying state. Instead, the agents must maintain a \textbf{Belief State} to represent its subjective understanding of the learning environment. 

\item $\mathcal{A}$ is the action space. An action $a \in \mathcal{A}$ involves content generation or the use of external tools, such as scheduling events via calendar APIs and issuing smart home control commands.

\item $T:\mathcal{S}\times\mathcal{A}\to P(\mathcal{S})$ is the state transition probability function. For each state-action pair $(s,a)$, it specifies a probability distribution over possible subsequent states. For example, executing a ``turn off lights" action may result in reaching the desired state with 90\% probability, while network latency or system errors could cause the state to remain unchanged with 10\% probability.

\item $\mathcal{G}$ is the goal space. Each goal $g \in \mathcal{G}$ specifies a particular objective of users. 

\item $\mathcal{O}$ is the observation space. Each observation $o \in \mathcal{O}$ represents a partial view of $s$, which can be manifested through different modalities. Partial observability may stem from multiple factors, such as the complexity of the learning environment and the limitations of the agent’s perceptual capabilities. 

\item $Z:\mathcal{S}\times\mathcal{A}\to \mathcal{O}$ is the observation function. For each state-action pair $(s,a)$, it provides a possible observation $o$ for the agent. 
    
\item $R:\mathcal{S}\times\mathcal{A}\to \mathcal{R}$ is the reward function, where $\mathcal{R}$ is the reward space. For each state-action pair $(s,a)$, it provides a feedback $r \in \mathcal{R}$, and $r$ is usually expressed in numerical form.
    
\item $\gamma \in [0,1)$ is the discount factor. It serves to balance the importance between immediate and future rewards. Since events that occur farther in the future are harder to predict accurately, it is generally desirable to reduce the significance of long-term rewards.    
    
\end{itemize}

\subsection{LLM-based Agent Loop}
\label{single_agent_loop}

Given a specific goal $g$, accomplishing it typically requires multiple execution loops. Furthermore, since the true environment state cannot be directly observed, an LLM-based agent needs to maintain a belief state $b_t$ that represents its internal and subjective understanding of the learning environment. In the multi-loop process, the agent continuously updates $b_t$, allowing $b_t$ to be dynamically refined across diverse contexts and over extended time spans. $b_t$ is the most fundamental unit in the agent's loop process, forming the basis of all agentic operations. Based on $b_t$, we further introduce the agent’s external behaviors including reasoning, execution, perception, and memorization. The concrete calculation in each loop can be given here:  

\begin{itemize}
    \item \textbf{Reasoning}: The agent first generates a plan $p_t$ for the next action conditioned on $b_t$ and $g$:
    \begin{equation}
    \label{reasoning_single_agent}
        p_t = \Phi(b_t, g).
    \end{equation}
    
    \item \textbf{Execution}: The agent then translates $p_t$ into an executable action $a_t$:
    \begin{equation}
    \label{excution_single_agent}
        a_t = E(b_t, p_t). 
    \end{equation}  
$a_t$ typically involves the invocation of relevant external tools.

\item \textbf{Feedback}: The learning environment would objectively provide a reward $r_t$ based on $s_t$ and $a_t$:
    \begin{equation}
        r_t = R(s_t, a_t).
    \end{equation}
    
\item \textbf{Environment Transition}: The learning environment transitions to the next state $s_{t+1} \in \mathcal{S}$ according to the following generated probability distribution:
    \begin{equation}
        {\rm Pr}(s_{t+1}|s_t, a_t) = T(s_t, a_t) .
    \end{equation}

\item \textbf{Perception}: The agent would then perceive $s_{t+1}$ and generate the observation $ o_{t+1} \in \mathcal{O}$:
\begin{equation}
\label{observation_single_agent}
    o_{t+1} = Z(s_{t+1}, b_t, a_t).
\end{equation}

    \item \textbf{Memorization}: Both $a_t$ and $o_{t+1}$ are used for updating its external memory module: 
        \begin{equation}
        \label{memory_single_agent}
            m_{t+1} = L_M(m_t, a_t, o_{t+1}).
        \end{equation}
    
    \item \textbf{Belief Update}: Based on the above decision-making process and feedback, the agent refines $b_{t}$ as follows,
        \begin{equation}
        \label{single_belief}
            b_{t+1} = L_B(b_t, m_{t+1}, a_t, r_t, o_{t+1}). 
        \end{equation}
 
\end{itemize}
The agent ultimately receives a cumulative discounted reward (i.e., $\sum_{t=0}^{\infty} \gamma^t r_t$) as a measure of goal achievement.

\par
Different from the single-agent setting, in the LLM-based MAS, each agent must communicate with other agents to accomplish goals. To model these processes, a communication structure $G_t$ is introduced among $N$ agents. Furthermore, $G_t$ would evolve with time to align with the dynamic adjustments of MAS. Therefore, compared with the above loop, the loop of LLM-based MAS is given as ``Reasoning--Execution--\textbf{Broadcasting}--Feedback--Environment Transition--Perception--Memorization--Belief Update--\textbf{Structure Evolution}". This loop includes two additional procedures: Broadcasting and Structure Evolution. Broadcasting means that the agent broadcasts its message to neighboring nodes according to its plan, while structure evolution indicates that the communication structure $G_t$ can be updated in each iteration. The complete description of the loop of MAS is given in Appendix A. 

\begin{tcolorbox}[colback=blue!5!white,colframe=black!75!white,
  title=\textbf{Definition of Agent hallucinations},
  fonttitle=\bfseries, coltitle=white, colbacktitle=black, rounded corners]
  
According to the proposed internal–external distinction, agent hallucinations refer to fictitious or erroneous operations in LLM-based agents, driven by overconfident ``human-like behaviors" in reasoning, execution, perception, memorization, and communication. Unlike LLM hallucinations, which are typically limited to linguistic errors, agent hallucinations go beyond language generation, manifesting as compound deviations arising from the coupling and interaction of multiple modules. These deviations can escalate to task failures and even physical safety risks. In contrast to other types of agent errors, agent hallucinations are often spontaneously generated within the agent’s own cognitive processes, representing an endogenous error form. Furthermore, owing to their cross-module propagation and hidden cognitive nature, agent hallucinations pose an especially serious threat, making them a central challenge in the pursuit of agent safety.
\end{tcolorbox}

\begin{figure*}[htbp]
    \centering
    \includegraphics[width=1\textwidth, height=11.5cm, trim=0cm 2cm 1cm 0cm, clip]{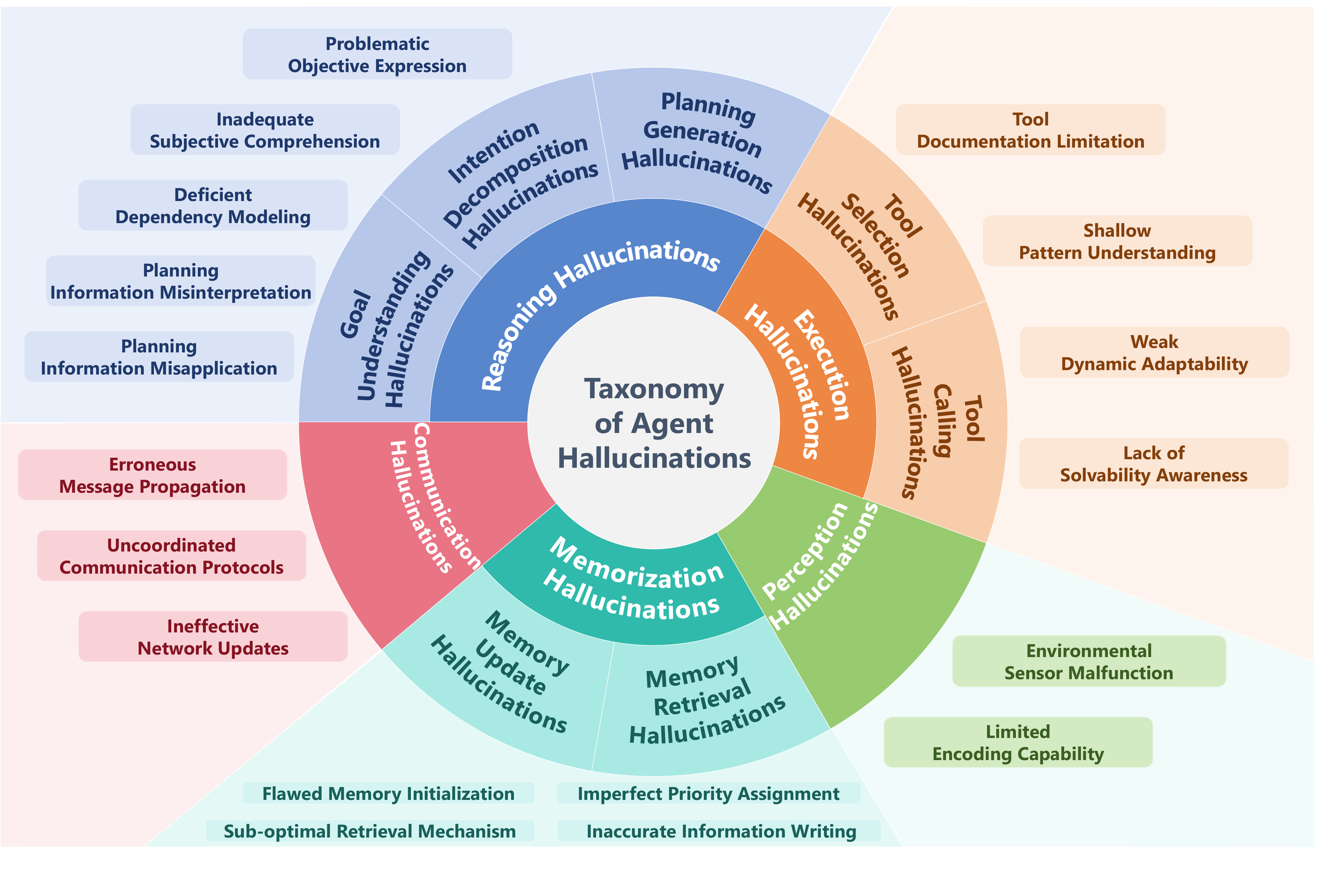} 
    \caption{A taxonomy of agent hallucinations. It includes five hallucination types and nine hallucination sub-types with corresponding triggering causes.}
    \label{fig:taxonomy}
\end{figure*}

\section{Taxonomy of Agent Hallucinations}
\label{section_3}

We first provide a formal definition of agent hallucinations. Based on this definition, we then introduce the new taxonomy that includes the following five types of agent hallucinations: Reasoning Hallucinations, Execution Hallucinations, Perception hallucinations, Memorization Hallucinations, and Communication Hallucinations, as illustrated in Fig.~\ref{fig:taxonomy}. Detailed descriptions of each type of agent hallucinations are provided in the following sub-sections. Representative hallucination examples are presented in Appendix B. 

\subsection{Reasoning Hallucinations}
Reasoning serves as the cornerstone of an agent's functionality, influencing behavior analysis and decision-making~\cite{cao2025large,he2025plan,zhang2025igniting}. Upon receiving a specific goal $g$, the agent first leverages its own reasoning capabilities to perform goal understanding for inferring the user's true intention: 

\begin{itemize}
\item \textbf{Goal Understanding}: This phase occurs before the agent executes multiple loops:    
    \begin{equation}
    \mathcal{I} = \text{Understand}(b_0,g), 
    \end{equation}
where $\mathcal{I}$ denotes the inferred intention, and $b_0$ represents the initial belief state. When $\mathcal{I}$ is complex and relatively difficult to execute, the agent would perform the intention decomposition to decompose $\mathcal{I}$ into a series of manageable sub-intentions $\{\mathcal{I}_t\}_{t=0}^n$. 

    \item \textbf{Intention Decomposition}: Typically, there are two decomposition methods: pre-defined decomposition and dynamic decomposition. In the first one, these sub-intentions are specified in advance:   
    \begin{equation}
    \label{a-decomposion}
    \{\mathcal{I}_t\}_{t=0}^{n} = \text{Pre-decompose}(b_0, \mathcal{I}).
    \end{equation}
    In the second one, the sub-intention for each loop is generated by 
    \begin{equation}
    \label{b-decomposion}
    \mathcal{I}_t = \text{Dyn-decompose}(b_t, \mathcal{I}).
    \end{equation}
The decomposition results are dynamically optimized based on the current belief state $b_t$, enabling timely and continuous refinement for the subsequent workflow~\cite{sapkota2025ai}. 
   
\item \textbf{Planning Generation}:
Each sub-intention $\mathcal{I}_t$ is then mapped to a concrete plan $p_t$ for the current loop:
    \begin{equation}
    \label{subplan_genereation}
    p_t = \text{Plan}(b_t, \mathcal{I}_t).
    \end{equation}
\end{itemize}

\begin{tcolorbox}[colback=blue!5!white,colframe=black!75!white,
  title=\textbf{Definition of Reasoning Hallucinations},
  fonttitle=\bfseries, coltitle=white, colbacktitle=black, rounded corners] 
Reasoning hallucinations refer to the phenomenons where LLM-based agents generate seemingly plausible plans which are in fact logically flawed or unsupported. Reasoning hallucinations occur in either of above three stages, and we categorize them into three types: \textbf{Goal Understanding Hallucinations (GUHs)}, manifested as misinterpretation of goal semantics, typically caused by problematic objective expression and inadequate subjective comprehension; \textbf{Intention Decomposition Hallucinations (IDHs)}, characterized by the flawed decomposition processes that produce irrelevant or infeasible sub-intentions, often due to deficient dependency modeling; \textbf{Planning Generation Hallucinations (PGHs)} represent hallucinations that occur when an agent generates each specific plan of a sub-intention, typically caused by the misinterpretation and misapplication of planning information.
\end{tcolorbox}

The subsequent sub-sections detail the triggering causes that give rise to these three types of reasoning hallucinations.

\noindent\emph{1) Problematic Objective Expression.} 
Understanding the user's true intention from $g$ is the first and essential step for effective  reasoning~\cite{zhang2025metamind}. Objectively, when the expression of goal information carries a certain degree of semantic vagueness, it can easily lead to erroneous parsing of user intention and induce reasoning hallucinations~\cite{zheng2023does,linok2024beyond}. This semantic vagueness can be mainly attributed to two issues: \textbf{Incomplete Goal Specification} and \textbf{Ambiguous Content Presence}~\cite{liu-etal-2023-afraid, keluskar2024llms}. To address semantic vagueness in user input, a plausible and effective approach is to endow the agent with the capability of active clarification~\cite{kobalczyk2025active,chen2025learning}. By engaging in appropriate proactive interactions with the user, the agent can gain a deeper understanding of the user's true needs and intentions.

\noindent\emph{2) Inadequate Subjective Comprehension.} 
In addition to above objective issues, there also exist subjective limitations in the agent’s comprehension capability. These are mainly manifested as \textbf{Instruction-following Deviation}~\cite{verma2025graft}, as well as \textbf{Long-range Contextual Misuse}~\cite{xu2025scrnet}. Specifically, instruction following refers to an agent's ability to adhere to user-provided instructions in order to accurately comprehend intentions~\cite{abbasian2023conversational}. When this procedure breaks down, the agent struggles to segment critical fields and extract key information from instructions, leading to distorted understanding and reasoning hallucinations~\cite{huang2024opera, wu2025clarifycoder, shi2024navigating}. Additionally, when agents handle excessively long input goals, they may fail to capture long-range contexts and instead over-rely on the most recent tokens, resulting in contextual misuse~\cite{zhao2023large, chu2024beamaggr, zhu2024knowagent}. Such deficiencies cause agents to violate explicitly stated user preferences~\cite{lai2024vision}.

\noindent\emph{3) Deficient Dependency Modeling.} 
Intention decomposition is a crucial phase where agents determine the successive reasoning process via the decomposed sub-intentions~\cite{wang2025medagent, tudor2025vecsr, zhao2025mactgmultiagentcollaborativethought}. These sub-intentions typically form a sequentially dependent chain, where the completion of each sub-intention pre-supposes the successful fulfillment of its predecessors. Inadequate modeling of dependency relationships among these sub-intentions can give rise to three types of errors: \textbf{Sub-intention Omission}, \textbf{Sub-intention Redundancy}~\cite{wan2024cot,li2024focus} and \textbf{Sub-intentions Disorder}. All three severely compromise reasoning integrity and efficiency, ultimately leading to reasoning hallucinations~\cite{wang2025tdag}. Sub-intention omission occurs when agents fail to identify essential sub-intentions, leading to the absence of critical reasoning steps~\cite{qiu2025co}. Sub-intention redundancy arises when agents erroneously introduce task-irrelevant sub-intentions, causing reasoning processes to deviate from the core intention and generating hallucinatory content~\cite{berijanian2025comparative, yu2025dyna, qiu2025co}. Sub-intention disorder refers to the phenomenon in which an agent arranges sub-intentions with a sequential relationship in the wrong order during reasoning, causing it to fail to obtain the results of prerequisite sub-intentions, thereby missing key information and further generating hallucinations~\cite{wang2023steps}.

\noindent\emph{4) Planning Information Misinterpretation.} 
In Eq.(\ref{subplan_genereation}), the agent relies on two key sources of information to generate sub-intention plans: \textbf{Operable Objects} and \textbf{Self-knowledge}. For operable objects, the agent must recognize both explicit and implicit relational connections between entities (e.g., spatial, temporal, or social relationships)~\cite{chu2025llm+, huang2024targa, zuo2025kg4diagnosis}. The misinterpreting information related to these operable objects can lead the agent to construct plans based on incorrect assumptions, thereby inducing planning generation hallucinations~\cite{manakul2023selfcheckgpt}. With respect to  self-knowledge, the agent is expected to plan reasonably within the bound of its own self-knowledge to avoid planning errors. When confronted with planning problems beyond its knowledge boundary, the agent tends to respond with excessive confidence, generating answers that sound certain but are actually incorrect~\cite{liu2025uncertainty}. 

\noindent\emph{5) Planning Information Misapplication.} 
Even when an agent correctly identifies the operable objects within its knowledge boundary, its ability to effectively utilize them remains constrained by its \textbf{Logical Reasoning} and \textbf{Introspection Capacity}~\cite{mirzadeh2024gsm}. Logical fallacies primarily manifest as rule conflicts and causal misattributions~\cite{liu2025enhancing}. Similar to humans, when agents encounter issues such as rule conflicts, causal inversion, or misattribution during causal reasoning~\cite{wu2025reasoning}, they often fail to complete the reasoning process or resolve underlying logical paradoxes, ultimately leading to erroneous planning~\cite{xu2025mragent}. Additionally, introspection is one of the key mechanisms by which agents achieve learning, cognition, and knowledge updating~\cite{shinn2023reflexion}, agents that lack sufficient introspection capacity are unable to effectively improve their performance by revisiting logical fallacies in their reasoning~\cite{belle2025agents}. Such agents may misinterpret feedback signals and accurately identify the sources of error~\cite{liang2023encouraging}, thereby making misguided adjustments to their reasoning processes~\cite{zhang2024self}. Even when feedback is correctly interpreted, the introspection process may still be flawed: agents might wrongly judge their reasoning as consistent despite existing logical contradictions, or mistakenly believe they have considered all relevant factors while overlooking critical variables~\cite{li2024hindsight}.

\begin{tcolorbox}[colback=blue!5!white,colframe=black!75!white,
  title=\textbf{Definition of Execution Hallucinations},
  fonttitle=\bfseries, coltitle=white, colbacktitle=black, rounded corners]
Execution hallucinations refer to the phenomenon where LLM-based agents claim to have completed certain sub-stages during the execution phase, but in reality, they have not actually been performed or accomplished. According to the above two sub-stages, we divide
execution hallucinations into two types: \textbf{Tool Selection Hallucinations (TSHs)}, where non-existent or irrelevant tools are chosen overconfidently; and \textbf{Tool Calling Hallucinations (TCHs)}, involving incorrect, omitted, or fabricated parameter fillings.
\end{tcolorbox}

\subsection{Execution Hallucinations}
The execution phase is where the agent translates its deliberated plan $p_t$ into a concrete, executable action $a_t$. $a_t$ typically involves invoking external tools, either as a single operation or through multiple parallel instances~\cite{zhu2025divide}. Broadly, this execution process in Eq.(\ref{excution_single_agent}) can be further decomposed into two sub-stages \cite{patil2024gorilla}:
\begin{itemize}
    \item \textbf{Tool Selection}: Given $p_t$ and a set of candidate tools $\mathcal{T}_\text{cand}$, the agent must first select the appropriate tool $T_s \in \mathcal{T}_\text{cand}$ as follows\footnote{$T_s$ may denote multiple tools that are invoked in parallel to accelerate execution.},
    \begin{equation}
        T_s=\text{Select} \left(b_t, p_t, \mathcal{T}_\text{cand} \right).
    \end{equation}
As there are a large number of tools, $\mathcal{T}_\text{cand}$ is typically retrieved from a full tool set $\mathcal{T}$ to narrow the selection scope~\cite{qin2023toolllm}:
    \begin{equation}
        \mathcal{T}_\text{cand}=\text{Retr} \left(b_t, p_t, \mathcal{T}\right).
    \end{equation}
    \item \textbf{Tool Calling}: Once $T_s$ is selected, the agent then needs to populate $T_s$ with the tool parameters derived from $b_t$ and $p_t$ to form the final executable action $a_t$:
    \begin{equation}
        a_t=\mathrm{Call} \left(b_t, p_t, T_s\right).
    \end{equation}
\end{itemize}

The underlying causes behind these two execution hallucinations are given in the following sub-sections.   

\noindent\emph{1) Tool Documentation Limitation.} 
Execution hallucinations often arise when the agent's internal representation of tool behaviors diverges from the actual functionality of tools, frequently due to limitation in \textbf{Tool Documentation}~\cite{r32}. These hallucinations occur because the agent believes it has correctly selected or invoked a tool, even though its decision is based on misleading documentation. Such deficiencies may include redundant information~\cite{r12, qu2025from}, incomplete or inaccurate descriptions~\cite{r13, r16}, or a lack of standardization~\cite{r14, r15}, all of which impair the agent's ability to properly use tools. 

\noindent\emph{2) Shallow Pattern Understanding.} 
 The agent's shallow understanding of \textbf{Tool Patterns} may lead to execution hallucinations wherein the agent confidently invokes invalid or outdated tools, mistakenly assuming successful execution. In fact, effective tool usage requires LLM-based agents to develop deep understanding of tool patterns. This requirement primarily stems from two key factors: First, tools often provide multiple functionalities that can be applied across diverse scenarios, rather than being limited to a single task-specific use case~\cite{r33}; Second, tools frequently exhibit complex collaborative patterns, such as sequential dependencies and the need for nested calls~\cite{r311}. However, LLM-based agents are typically trained with insufficient exposure to diverse and complex tool-use scenarios~\cite{r34, r35, r36}. This limitation makes agents prone to hallucinating tool invocations that may appear plausible or confident, but in fact violate expected patterns or omit essential details, especially when dealing with complex and novel tasks~\cite{r22}.

\noindent\emph{3) Weak Dynamic Adaptability.} 
LLM-based agents are typically trained on relatively static datasets, with their tool-use knowledge embedded in fixed model parameters~\cite{r35,r36}. However, the execution environment of tools is inherently dynamic and continuously evolving. Tool functionalities may evolve, API interfaces may be modified, and in some cases, tools may be deprecated or replaced~\cite{r22,lumer2025scalemcp,xian2025measuring}. When an agent lacks sufficient adaptability to these \textbf{Tool Dynamics}, its tool-use behavior becomes misaligned with the actual environment, thus leading to execution hallucinations wherein the agent confidently invokes outdated or invalid tools while incorrectly assuming successful execution.

\noindent\emph{4) Lack of Solvability Awareness.} 
\textbf{Tool Solvability} refers to whether the current plan $p_t$ can be successfully executed under existing conditions. It is primarily related to two factors: a) the availability of suitable tools in $\mathcal{T}$; b) the clarity and completeness of $p_t$. A lack of solvability awareness in LLM-based agents can also lead to execution hallucinations, where the agent mistakenly assumes that $p_t$ is solvable and proceeds with unjustified confidence. For example, if no suitable tools are available, any tool retrieved from $\mathcal{T}$ is likely to be irrelevant or even fabricated and non-executable, resulting in tool selection hallucinations~\cite{r54}. Furthermore, even when a suitable tool exists and is correctly selected, an unclear or incomplete plan $p_t$ can prevent the agent from correctly deriving the required parameters for tool invocation~\cite{wang2024learning}. Consequently, the agent may omit required parameters or fabricate parameter values, falsely believing that the tool call is valid.

\subsection{Perception Hallucinations}

Analogous to how humans depend on sensory organs like eyes and ears to acquire information from the external world and convert it into neural signals, LLM-based agents also rely on a perception module to facilitate interaction with the learning environment. The perception module serves as a fundamental interface, extending its perception into a multi-modal space that includes textual, auditory, and visual modalities. As illustrated in Eq.(\ref{observation_single_agent}), the agent receives the external information (e.g., $s_{t+1}$) and transforms this information into internal observations (e.g., $o_{t+1}$) via the perception module. Here we analyze the causes of perception hallucinations.

\begin{tcolorbox}[colback=blue!5!white,colframe=black!75!white,
  title=\textbf{Definition of Perception Hallucinations},
  fonttitle=\bfseries, coltitle=white, colbacktitle=black, rounded corners]
Perception hallucinations refer to the phenomenon where, during the process of receiving and transforming external information, an agent produces internal observations that significantly deviate from or contain factual errors about the actual learning environment. This typically arises from environmental sensor malfunction or limited encoding capability. Such hallucinations undermine the agent’s accurate perception and further disrupt subsequent decision-making processes.
\end{tcolorbox} 

\noindent\emph{1) Environmental Sensor Malfunction.} 
Agents typically rely on environmental sensors to collect external data and convert it into digital signals~\cite{trivedi2025intelligent}. Common sensor types include visual sensors (e.g., cameras), auditory sensors (e.g., microphones), tactile sensors (e.g., pressure-sensitive pads), and inertial measurement units (e.g., accelerometers)~\cite{liu2025advances}. When environmental sensor malfunctions occur, such as lens distortion in cameras or signal drift in inertial measurement units, agents may fail to accurately receive raw input information, potentially resulting in perception hallucinations. 
 
\noindent\emph{2) Limited Encoding Capability.}
After receiving \( s_{t+1} \), LLM-based agents typically employ a specific encoding module to generate \(o_{t+1}\), which serves as a high-level representation of the received signals such as images and sound waves. However, this encoding process encounters two critical limitations: \textbf{a) Insufficient Unimodal Representation}: Agents struggle to extract the key information of individual modality from \( s_{t+1} \). This limitation arises from multiple factors, potentially including insufficient training data quality~\cite{yu2024hallucidoctor}, the Transformer architecture’s tendency~\cite{vaswani2017attention} to overlook local details~\cite{chen2024florence}, and training conflicts between the pre-training and fine-tuning stages~\cite{liu2025visual}. \textbf{b) Weak Cross-modal Collaboration}: Agents lack an effective mechanism to integrate semantic associations across different modalities, resulting in the incomplete representation of \( s_{t+1} \). When faced with a complex learning environment, LLM-based agents typically perform the procedures of modality alignment and joint encoding to integrate different multi-modal information~\cite{ghazanfari2024emma}. However, during this process, LLM-based agents may fail to align modalities due to insufficient cross-modal annotations and imbalanced modality data~\cite{lai2024lisa}. Furthermore, in the generation stage, agents may be dominated by the language prior, gradually neglecting non-textual information~\cite{wang2024vigc}.

\begin{tcolorbox}[colback=blue!5!white,colframe=black!75!white,
  title=\textbf{Definition of Memorization Hallucinations},
  fonttitle=\bfseries, coltitle=white, colbacktitle=black, rounded corners]
Memorization hallucinations refer to the phenomenon in which an agent implicitly assumes its memory to be accurate and reliable, without validating the correctness of the stored content. Consequently, when the agent incorporates its memory module into decision-making, it may rely on outdated, fabricated, or conflated memories, resulting in flawed reasoning or inappropriate actions. Based on the above two operations, we categorize memorization hallucinations into two types: \textbf{Memory Retrieval Hallucinations (MRHs)}, where irrelevant or non-existent information are retrieved; and \textbf{Memory Update Hallucinations (MUHs)}, involving incorrect modifications and deletions of memory content.
\end{tcolorbox}

\subsection{Memorization Hallucinations}

The memory module is a core component of LLM-based agents, tasked with storing and managing information to facilitate subsequent decision-making~\cite{xiong2025memorymanagementimpactsllm, packer2023memgpt,huang2023memory}. It primarily owns two operations:  \textbf{Memory Retrieval}: Extracting and integrating relevant information from stored memory to support the current decision process; \textbf{Memory Update}: revising and removing existing memory based on newly acquired information or feedback to ensure its accuracy and timeliness.
    
Subsequently, we analyze the underlying causes that trigger the above two memorization hallucinations.

\noindent\emph{1) Flawed Memory Initialization.}
The \textbf{Initial Memory} $m_0$ serves as the foundation of memorization in LLM-based agents, and its quality directly influences the reliability of subsequent memory processes. Biased or incomplete content in \( m_0 \) can introduce initialization deficiencies, leading to memorization hallucinations before deployment. In particular, certain biases, particularly those related to gender and nationality, are inherently linked to hallucination issues~\cite{huang2025survey}. Moreover, when critical information is missing (e.g., the time or location of a historical event) in $m_0$, agents may rely on inherent biases or assumptions to fill in the gaps, resulting in inaccurate or fabricated responses in many real-world scenarios~\cite{li2024banishing,singh2025agentic,singhal2025toward}.

\noindent\emph{2) Sub-optimal Retrieval Mechanism.} First, poor \textbf{Ranking Strategies} may lead the agent to retrieve memory content that only superficially "appears similar" but lacks true relevance~\cite{tu2025rbft,Agrawal2024MindfulRAGAS}. Additionally, deficiencies in the memory \textbf{Indexing Structure}~\cite{he2024g,Modarressi2023RETLLMTA}, such as inappropriate index granularity or delayed index updates, can further exacerbate this problem, potentially resulting in information loss and the retrieval of outdated memory. Finally, insufficient understanding of \textbf{Query Semantics} is also an important problem. If the agent misinterprets the query’s intent or overlooks specific task requirements~\cite{zhang2025hallucination,yoon2025ask}, it may fail to capture contextual cues necessary for accurate retrieval.

\noindent\emph{3) Imperfect Priority Assignment.}
This issue mainly manifests in two aspects: \textbf{Forgetting Priority} and \textbf{Merging Priority}. For memory forgetting, poorly assigned priorities can result in the elimination of important information or the retention of irrelevant content, ultimately compromising the accuracy of subsequent decisions~\cite{packer2023memgpt,mei2025survey}. Moreover, when the agent merges multiple memory fragments, failure to correctly assess their priorities~\cite{jin2024tug,ding2024retrieve} may result in the merged memory containing inherent conflicts~\cite{wang2025accommodate,wu2024faithful}. Some of these conflicts are implicit~\cite{xu2024knowledge}: Although memory fragments may appear similar, they differ significantly in key semantic details, thereby increasing the difficulty of priority determination.

\noindent\emph{4) Inaccurate Information Writing.}
When performing long-term tasks and engaging in multi-turn interactions, agents must summarize, structure, and store relevant historical information to the memory module. However, this process is susceptible to \textbf{Information Compression} issues, where the generated summaries may be overly general, omit crucial details~\cite{xu2025mem,qin2025towards,wang2025recursively}, or introduce distortions due to imperfect abstraction~\cite{Guo2025RepoAuditAA,ye2025task,xiong2025memorymanagementimpactsllm}. Additionally, non-standardized \textbf{Memory Formats} with disorganized structures can hinder writing efficiency~\cite{xiong2025memorymanagementimpactsllm,wang2025mirix}. Memory \textbf{Capacity Constraints} exacerbate these challenges, because limited storage may necessitate the selective retention of information~\cite{packer2023memgpt,zhang2024survey}, increasing the risk that salient but less frequently accessed information is discarded.  

\subsection{Communication Hallucinations}

\begin{tcolorbox}[colback=blue!5!white,colframe=black!75!white,
  title=\textbf{Definition of Communication Hallucinations},
  fonttitle=\bfseries, coltitle=white, colbacktitle=black, rounded corners]

Communication hallucinations refer to the phenomenon in which LLM-based agents appear to engage in meaningful inter-agent communication, yet the information exchanged is inaccurate, misleading, or fabricated, thereby undermining collaboration. Such hallucinations typically arise from erroneous message propagation, where incorrect information is shared as if reliable; uncoordinated communication protocols, where inconsistent formats or timing lead to misunderstandings; and ineffective network updates, where outdated or poorly synchronized connections distort information flow and hinder task execution.
\end{tcolorbox}

Compared with the independent operations of a single agent, LLM-based MAS emphasizes collaboration and coordination to harness collective intelligence for solving more complex and demanding tasks~\cite{tran2025multi}. In such systems, effective inter-agent communication serves as an essential requirement~\cite{yan2025beyond}, facilitating the exchange of ideas and the coordination of plans among agents~\cite{piatti2024cooperate}. 

The following sub-sections provide a detailed account of triggering causes underlying communication hallucinations.

\noindent\emph{1) Erroneous Message Propagation.} 
In LLM-based MAS, agents fundamentally rely on LLMs~\cite{xi2025rise, cheng2024exploring,liu2025advances} to generate messages for exchanging information with other agents. However, since LLMs are prone to the well-known \textbf{Factuality and Faithfulness Hallucinations}~\cite{huang2025survey}, some agents may produce messages containing inaccurate facts, misinterpretations of shared knowledge, or misleading inferences~\cite{yoffe2024debunc,cemri2025multi,zhang2025agent}, thereby giving rise to communication hallucinations. Beyond this, \textbf{Content Redundancy} is also an important cause, where agents generate unnecessary or repetitive content that obscures critical signals, increases cognitive load, and sometimes leads to redundant task execution steps that manifest as logical errors~\cite{zhang2025cut, wang2025agentdropout, cemri2025multi}. \textbf{Information Asymmetry} can further exacerbate this issue. Because agents own different roles and positions within MAS, the information accessible to them varies, and such asymmetric settings may yield vague or incomplete instructions that hinder
task comprehension~\cite{liu2024autonomous} and amplify the risk of biased decisions~\cite{wang2025learning}.

\noindent\emph{2) Uncoordinated Communication Protocols.} 
Communication protocols govern how agents exchange messages, directly determining the efficiency, reliability, and coordination of their interactions~\cite{ehtesham2025survey, yan2025beyond, yang2025survey}. Without a unified and effective protocol, agents may "talk past each other", leading to communication hallucinations. First, LLM-based MAS usually follows a manner of \textbf{Asynchronous Scheduling}~\cite{hong2024metagpt}, so when receiving and processing instructions, agents may encounter issues of information loss and information overload. Such temporal discrepancies can result in information errors, thereby increasing the risk of hallucinatory outputs~\cite{gim2024asynchronous, gonzalez-pumariega2025robotouille}. Second, communication protocols define the \textbf{Message formats}. Current LLM-based agents predominantly rely on the format of natural language~\cite{bansal2024challenges} which often introduces instruction ambiguity. Adopting structured formats (e.g., JSON) can improve clarity and rigor of expression, which mitigates the risk of miscommunication~\cite{yang2024multi, wang2025talk}. Finally, LLM-based MAS demands a robust \textbf{Fault-tolerant Design}, incorporating confirmation conditions and synchronization constraints to avoid erroneous decisions caused by message loss or delays in dynamic or noisy environments~\cite{liu2024a,kwon2025cp}.

\noindent\emph{3) Ineffective Network Updates.} Network topology defines how agents are interconnected, determining who communicates with whom and how frequently \cite{yan2025beyond,park2023generative,chen2024agentverse}. As discussed in Appendix A, the network topology is continuously evolving, and network updates reshape the propagation paths of messages within MAS. When network updates are ineffective, they can induce communication hallucinations due to inconsistent or outdated inter-agent connections~\cite{cemri2025multi, zhang2025agent, zhuge2024gptswarm}. Although recently proposed strategies improve the flexibility and responsiveness of MAS, they often suffer from delayed updates and poor coordination~\cite{park2023generative, song2025code, aratchige2025llms}. If the updated network fails to accurately reflect agents’ current relevance or expertise, messages may be routed to inappropriate recipients, leading to misunderstandings or redundant reasoning in LLM-based MAS.

\section{Systematic Methodology}
\label{Systematic_Methodology}
\begin{figure*}[htbp]
    \centering
    \includegraphics[width=1\textwidth, trim=0cm 0cm 0cm 0cm, clip]{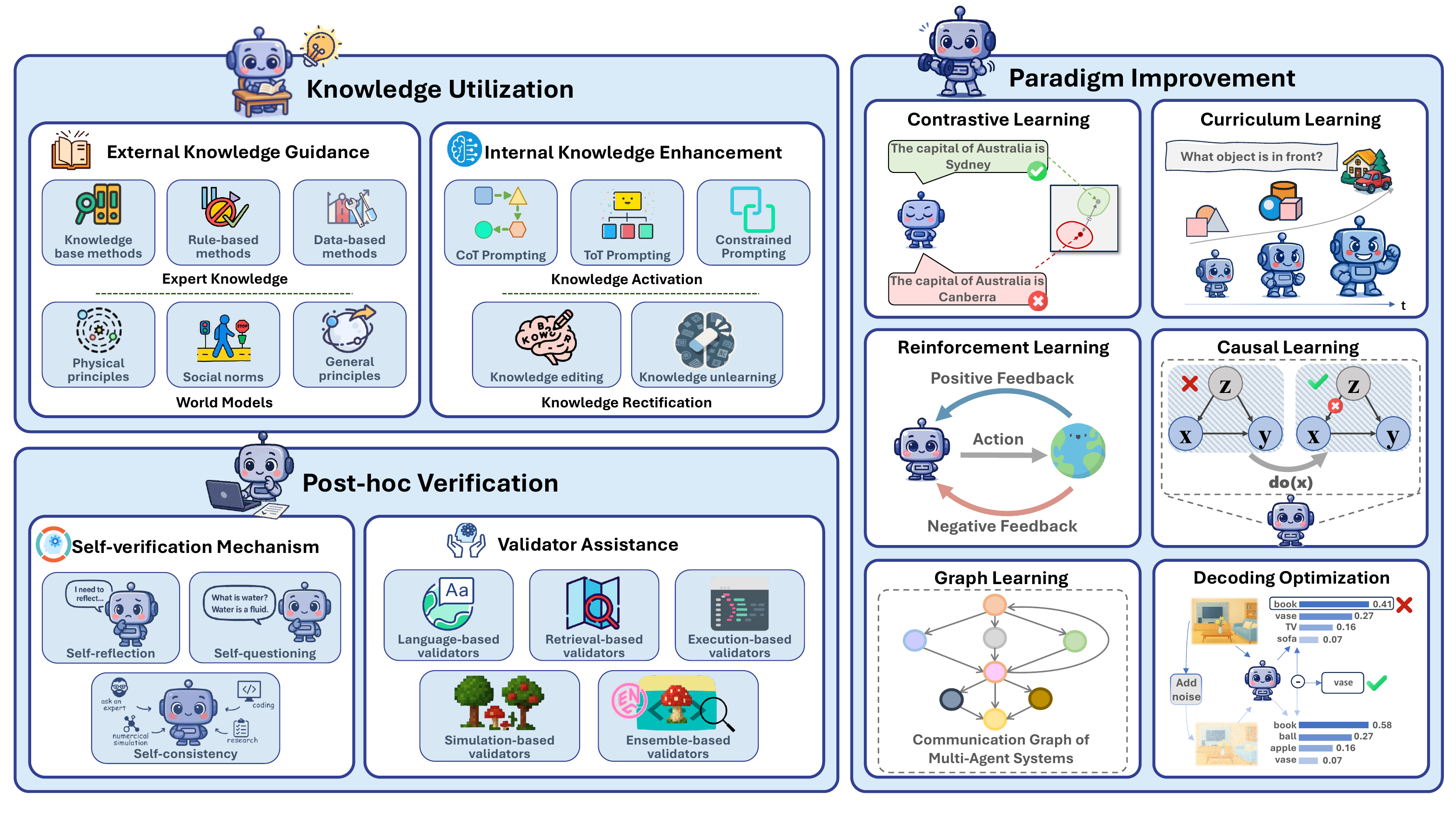} 
    \caption{A simple illustration of approaches to agent hallucination mitigation. It encompasses three branches, knowledge utilization, paradigm improvement, and post-hoc certification, comprising a total of ten representative methods.}
    \label{fig:Agent Hallucination Mitigation}
\end{figure*}

In this section, we provide a comprehensive review of recent methods for mitigating and detecting hallucinations in agents. For mitigation, we systematically group existing approaches into three main categories in Section~\ref{section_4_1}: \textbf{1) Knowledge Utilization}, which leverages external and internal knowledge to address hallucination issues; \textbf{2) Paradigm Improvement}, which focuses on developing advanced learning paradigms to prevent agent hallucinations during training and inference; and \textbf{3) Post-hoc Verification}, which introduces verification techniques to calibrate agent outputs and reduce the risk of hallucinations. In Fig.~\ref{fig:Agent Hallucination Mitigation}, we provide a simple illustration of each specific mitigation approach. For detection, existing methods are fewer compared to mitigation, so we place detection methods after mitigation methods. In Section~\ref{section_4_2}, we review the available detection approaches corresponding to each type of agent hallucinations.

\subsection{Agent Hallucination Mitigation}
\label{section_4_1}

\noindent\emph{1) Knowledge Utilization.} This category equips LLM-based agents with accurate and reliable information support, enabling multi-dimensional regulation of their reasoning, execution, perception, and memorization behaviors. This, in turn, helps reduce hallucinatory outputs stemming from knowledge gaps and biases. Here we categorize knowledge utilization into two types: \textbf{a) External Knowledge Guidance}: Agent hallucinations often arise from insufficient knowledge guidance. Therefore, external knowledge serves as a critical resource for mitigating hallucinations and enhancing output reliability. \textbf{b) Internal Knowledge Enhancement}: This line of research focuses on activating or rectifying the agent’s internal
knowledge to better mitigate hallucinations.

\noindent\emph{a) External Knowledge Guidance (EKG)\footnote{The abbreviation is added here mainly to save space in Table~\ref{table:mitigation}.}.} We categorize the existing approaches of external knowledge guidance into two types: \textbf{Expert Knowledge} and \textbf{World Models}. 

Expert knowledge refers to the domain-specific information distilled from the behaviors and decision-making processes of human experts or high-performing systems during task execution. Owing to its high reliability and precision, expert knowledge serves as a robust external reference, significantly reducing the likelihood of hallucinations across various agentic operations. In related literature, expert knowledge can be manifested different forms, and we list several representative forms:  \textbf{i) Knowledge base methods}: Expert knowledge can be stored as knowledge bases, such as introspective reasoning knowledge bases~\cite{liang2024introspective}, action knowledge bases~\cite{zhu2024knowagent}, experiential knowledge bases~\cite{fu2025knowmap}, from which agents can retrieve real cases and incorporate them into prompts to support more accurate decision-making~\cite{wang2025medagent}. \textbf{ii) Rule-based methods}: expert knowledge can be leveraged to establish rules such as mathematical rules~\cite{ liu2025enhancing}, linguistic rules~\cite{chu2025llm+}, procedural rules~\cite{yuan2024easytool} or MAS communication protocols~\cite{kwon2025cp}. These rules can explicitly constrain the agent's external behaviors, mitigating agent hallucination risks~\cite{hsieh2023tool}. \textbf{iii) Data construction methods}: By leveraging expert annotations and prompting techniques, synthetic datasets can include diverse and targeted instruction–response pairs~\cite{abdelaziz2024granite}, Q\&A pairs~\cite{dong2025insight} or preference pairs~\cite{huang2025enhance}. These curated datasets can enhance the agent's capability to handle complex scenarios, including fine-grained reasoning, tool usage, and multi-modal perception. 
\par
World models refer to systems that construct useful representations of the world, equipping agents with foundational world knowledge~\cite{long2025survey}. This enables them to handle real-world problems and reject outputs that contradict established facts, thereby effectively mitigating the occurrence of agent hallucinations~\cite{richens2025general}. In POMDP setting, world models are particularly important for constructing more accurate estimations of partially observable environments~\cite{hu2023language,zhang2024combo}. The scope of world knowledge encompasses a wide range of domains, from fundamental \textbf{Physical Principles}~\cite{duan2024enhancing} such as Newton’s laws of motion and the conservation of mass to complex \textbf{Social Systems and Structures}~\cite{zhao2023large, chae2024web}. For example, NavMorph~\cite{yao2025navmorph} leverages the world model to provide knowledge of the physical environment, avoiding navigation biases caused by environmental unfamiliarity, while WKM~\cite{qiao2024agent} supplies common-sense knowledge—such as the need to clean ingredients before cooking—to achieve behaviors aligned with human expectations.
\par
\noindent\emph{b) Internal Knowledge Enhancement (IKE).} Current enhancement strategies can be broadly classified into two types: \textbf{Internal Knowledge Activation} and \textbf{Internal Knowledge Rectification}. The former aims to leverage \textbf{Prompt Engineering} to carefully stimulate and utilize the agent's internal knowledge, guiding it to fully exploit its capabilities. Here we highlight several representative prompting techniques: \textbf{i) Chain-of-Thought (CoT)}: By incorporating step-by-step reasoning instructions~\cite{wei2022chain,manas2024cot} into the prompt, CoT guides the agent to break down complex problems and output intermediate steps in a logical sequence, thereby enhancing the accuracy and reliability of handling multi-step tasks. \textbf{ii) Tree-of-Thought (ToT)}: Building upon CoT, ToT explores and evaluates multiple reasoning paths in parallel~\cite{wang2022self,chu2024beamaggr}, fully utilizing the diversified information within the agent's internal knowledge to strengthen the completeness and robustness of its reasoning process.  \textbf{iii) Constrained Prompting}: By incorporating various forms of constraints~\cite{singh2023progprompt,goncharov2025segment} such as semantic~\cite{chen2024refining} and spatial constraints~\cite{wu2025mitigating} into prompts, this approach guides the agent to focus on task-relevant information, effectively reducing the generation of irrelevant content.
\par
Internal knowledge rectification generally involves two representative paradigms. The first is \textbf{Knowledge Editing} which replaces inaccurate or outdated knowledge with correct knowledge while minimizing disruptions to other internal knowledge~\cite{li2024banishing}. It is typically divided into two main approaches~\cite{yao2023editing}: the locate-and-edit approach~\cite{Li2023PMETPM}, which identifies and modifies specific knowledge components; the meta-learning approach~\cite{de2021editing}, which aims to to learn how to update the model's parameters for adjusting knowledge for new information. The second is \textbf{Knowledge Unlearning}. In contrast to knowledge editing, it focuses on  removing erroneous knowledge to improve the agent’s reliability. It includes two main approaches~\cite{si2023knowledge}: parameter optimization~\cite{sun2025unlearning}, which fine-tunes model parameters under certain conditions (e.g., parameter update scopes\cite{chen2023unlearn}) to forget some targeted knowledge; parameter merging~\cite{ilharcoediting}, which performs offline operations on model parameters without further training.

\noindent\emph{2) Paradigm Improvement.} 
This section focuses on enhancing the agent’s capabilities through general learning paradigms to mitigate hallucinations. We summarize six representative paradigms as follows. 

\noindent\emph{a) Contrastive Learning (CoL)} is a well-established self-supervised learning paradigm whose core idea is to learn more discriminative representations by comparing similarities and differences between samples~\cite{le2020contrastive,cao2021bipartite,lin2025contrastive}. By incorporating contrastive learning, the agent can better identify relevant and accurate patterns, reducing errors when handling unfamiliar or incomplete task inputs and effectively mitigating hallucination issues~\cite{chia2023contrastive,li2025refine,jiang2024hallucination}. 

\noindent\emph{b) Curriculum Learning (CuL)} is a learning paradigm inspired by the way humans learn. Its core idea is to start training the model with "easy" examples and gradually move to "harder" ones, improving learning efficiency and generalization ability~\cite{le2020contrastive}. By leveraging curriculum learning, agents rapidly accumulate successful experiences by starting with simpler tasks and gradually transferring this knowledge to more complex scenarios. This progressive learning paradigm enhances their fundamental capabilities, effectively mitigating agent hallucinations~\cite{deng2024can, xi2024training,r33}.
\par
\noindent\emph{c) Reinforcement Learning (RL)} serves as a fundamental learning paradigm, emulating the biological learning process of "trial, feedback, and adjustment". This approach enables agents to interact with their environment and learn to maximize cumulative rewards through iterative trial and error~\cite{kaelbling1996reinforcement}. Through the reinforcement learning paradigm, the agent can adjust its policy based on rewards, thereby optimizing future decision-making, learning to achieve goals in complex environments, and significantly mitigating potential hallucination issues in multi-turn interactions~\cite{xiong2025mpo,shen2025satori,r43,ben2023mocha,yu2024rlhf}. In addition, there are also studies that aim to improve the reliability of reward signals in order to enhance the overall model effectiveness~\cite{zhang2024longreward}. 

\noindent\emph{d) Causal Learning (CaL)} refers to the process in statistical machine learning that goes beyond capturing mere correlations and instead aims to model, discover, and leverage causal relationships between variables, thereby enhancing the model’s interpretability and generalization ability~\cite{pearl2009causal,lin2025generative}. In the context of LLM-based agents, identifying and leveraging potential causal relationships within task inputs and decision-making processes enables more accurate understanding of how different factors influence task outcomes. This helps reduce representational errors and logical inconsistencies for mitigating perception~\cite{li2025treble,hu2025causal} and reasoning hallucinations~\cite{wang2024csce,tang2023towards}. 

\definecolor{light-green}{RGB}{240, 247, 236}
\definecolor{light-blue}{RGB}{237, 242, 248}
\definecolor{light-gray}{RGB}{247, 247, 247}
\definecolor{light-yellow}{RGB}{252, 248, 232}
\definecolor{light-orange}{RGB}{252, 241, 232}
\definecolor{Palette}{HTML}{7A7E7D}

\begin{table*}[ht]
\caption{An overview of mitigation methods. To save space, we use the abbreviations of hallucination types and mitigation methods mentioned earlier. ``\ding{52}" indicates that related works already exist; ``--" indicates that it has not yet been explored.} 
\label{table:mitigation}
\centering
\setlength{\tabcolsep}{8pt} 
\renewcommand{\arraystretch}{1.2} 
\newcommand{\g}{\cellcolor{green!10}}

\begin{tabular}{c|c|c|c|c|c|c|c|c|c|c|c} \toprule
\multicolumn{2}{c|}{\multirow{2}{*}{\diagbox{Type}{Branch}}}
& \multicolumn{2}{c|}{\makecell{Knowledge\\Utilization}} & \multicolumn{6}{c|}{\makecell{Paradigm Improvement}} & \multicolumn{2}{c}{\makecell{Post-hoc\\Verification}} \\ \cmidrule(lr){3-12} 
 \multicolumn{2}{c|}{}& EKG & IKE & CoL & CuL & RL & CaL & GL & DO & SM & VA \\ \midrule
 
\rowcolor{light-green}
& GUHs & \textcolor{ForestGreen}{\ding{52}} & \textcolor{ForestGreen}{\ding{52}} &\textcolor{ForestGreen}{\ding{52}} &-- &\textcolor{ForestGreen}{\ding{52}} &-- &-- &-- & \textcolor{ForestGreen}{\ding{52}} & \textcolor{ForestGreen}{\ding{52}} \\
\rowcolor{light-green}
& IDHs & \textcolor{ForestGreen}{\ding{52}} & \textcolor{ForestGreen}{\ding{52}} &-- &-- &-- &-- & \textcolor{ForestGreen}{\ding{52}}& \textcolor{ForestGreen}{\ding{52}}& \textcolor{ForestGreen}{\ding{52}} & \textcolor{ForestGreen}{\ding{52}} \\ 
\rowcolor{light-green}
\multirow{-3}{*}{\makecell{Reasoning\\Hallucinations}}
& PGHs & \textcolor{ForestGreen}{\ding{52}} & \textcolor{ForestGreen}{\ding{52}} &\textcolor{ForestGreen}{\ding{52}} &\textcolor{ForestGreen}{\ding{52}} & \textcolor{ForestGreen}{\ding{52}} & \textcolor{ForestGreen}{\ding{52}} & \textcolor{ForestGreen}{\ding{52}}&\textcolor{ForestGreen}{\ding{52}} & \textcolor{ForestGreen}{\ding{52}} & \textcolor{ForestGreen}{\ding{52}} \\ \hline

\rowcolor{light-blue}
& TSHs & \textcolor{RoyalBlue}{\ding{52}} & \textcolor{RoyalBlue}{\ding{52}} & \textcolor{RoyalBlue}{\ding{52}} &\textcolor{RoyalBlue}{\ding{52}} & \textcolor{RoyalBlue}{\ding{52}} &--  & \textcolor{RoyalBlue}{\ding{52}} &-- & \textcolor{RoyalBlue}{\ding{52}} & \textcolor{RoyalBlue}{\ding{52}} \\
\rowcolor{light-blue}
\multirow{-2}{*}{\makecell{Execution\\Hallucinations}}
& TCHs & \textcolor{RoyalBlue}{\ding{52}} & \textcolor{RoyalBlue}{\ding{52}} & \textcolor{RoyalBlue}{\ding{52}} &\textcolor{RoyalBlue}{\ding{52}} & \textcolor{RoyalBlue}{\ding{52}} &-- &--  & --& \textcolor{RoyalBlue}{\ding{52}} & \textcolor{RoyalBlue}{\ding{52}} \\  \hline

\rowcolor{light-gray}
\multicolumn{2}{c|}{\cellcolor{light-gray}\colorbox{light-gray}{\makecell{Perception\\Hallucinations}}}
& \textcolor{Palette}{\ding{52}} & \textcolor{Palette}{\ding{52}} &\textcolor{Palette}{\ding{52}} &-- & \textcolor{Palette}{\ding{52}} & \textcolor{Palette}{\ding{52}} &-- & \textcolor{Palette}{\ding{52}} & -- & \textcolor{Palette}{\ding{52}} \\  \hline

\rowcolor{light-yellow}
& MRHs & \textcolor{YellowOrange}{\ding{52}} &-- &-- &-- &--  &-- & \textcolor{YellowOrange}{\ding{52}} & --& -- & \textcolor{YellowOrange}{\ding{52}} \\
\rowcolor{light-yellow}
\multirow{-2}{*}{\makecell{Memorization\\Hallucinations}}
& MUHs & \textcolor{YellowOrange}{\ding{52}} & \textcolor{YellowOrange}{\ding{52}} &\textcolor{YellowOrange}{\ding{52}} &-- &--  &-- & \textcolor{YellowOrange}{\ding{52}} & --& -- & \textcolor{YellowOrange}{\ding{52}} \\ \hline

\rowcolor{light-orange}
\multicolumn{2}{c|}{\cellcolor{light-orange}\colorbox{light-orange}{\makecell{Communication\\Hallucinations}}}
& \textcolor{BurntOrange}{\ding{52}} &-- &-- &-- & \textcolor{BurntOrange}{\ding{52}} & -- & \textcolor{BurntOrange}{\ding{52}} &-- & \textcolor{BurntOrange}{\ding{52}} & \textcolor{BurntOrange}{\ding{52}} \\ \bottomrule
\end{tabular}

\end{table*}

\noindent\emph{e) Graph Learning (GL)} is a class of machine learning paradigms specifically designed to handle graph-structured data, which has been widely applied in various domains such as social networks, bioinformatics, and recommendation systems~\cite{xia2021graph,lin2024graph,zhu2025ttgl}. It enhances the agent’s capability to organize and manage tasks in a structured and systematic manner, particularly for tool usage and memory updates, thereby reducing the risk of agent hallucinations~\cite{bei2025graphs, wu2024graph,liu2025graph}. For example, ControlLLM~\cite{2023controlllm} builds a tool graph comprising resource and tool nodes, and employs a parallel graph search algorithm to efficiently find appropriate tool paths for mitigating execution hallucinations. More importantly, recent studies leverage graph learning to optimize communication topology in LLM-based MAS~\cite{zhuge2024gptswarm, yang2025agentnet, chen2024internet}. By dynamically learning which agents should communicate, when, and through what channels, the system can suppress noisy message passing while promoting informative interactions. 
\par 
Unlike the aforementioned training-phase learning paradigms, \emph{f) Decoding Optimization (DO)} is a test-time learning paradigm. This paradigm can improve output quality by adjusting probability distributions or attention patterns in decoding processes, ensuring better alignment with inputs and factual knowledge,thereby reducing hallucinations stemming from reasoning errors~\cite{huang2024opera,wang2024mllm,tang2025seeing}. Common decoding strategies include contrastive decoding, self-calibrated attention decoding~\cite{tang2025intervening} and fusion decoding~\cite{wang2024mllm}. Contrastive decoding~\cite{leng2024mitigating} selects more reliable and context-consistent outputs by comparing multiple decoding candidates. Self-calibrated attention decoding~\cite{tang2025intervening} aims to dynamically adjust the attention weights to capture more informative cues. Fusion decoding incorporates the multi-source information to reduce dependence on single-modal priors for mitigating hallucinations. CGD~\cite{deng2024seeing} is one of representative works, which integrates semantic similarity scores between textual and visual content with final-layer token likelihoods to reduce reliance on language priors.

\noindent\emph{3) Post-hoc Verification.}
This paradigm focuses on monitoring and evaluating outputs after task execution. It generally follows a multi-step process in which, immediately after each step, the agent assesses the validity, consistency, and factuality of the intermediate decisions and actions generated in that specific step. This assessment helps prevent the accumulation and propagation of hallucinations in long-horizon tasks. Existing approaches can be broadly categorized into two types: \textbf{a) Self-verification Mechanism}, in which the agent introspectively reviews its own behaviors using an internal reasoning strategy or prompt-based self-assessment. \textbf{b) Validator Assistance}, which leverages another independent system to detect potential flaws in the agent’s outputs.
  
\noindent\emph{a) Self-verification Mechanism (SM).} Self-verification is a lightweight and model-internal approach wherein agents assess the validity and reliability of their own outputs without relying on external validators. It plays a crucial role in hallucination mitigation~\cite{moncada2025agentic}, and several technical methods have been explored to implement self-verification. Specifically, \textbf{Self-reflection} enables agents to revisit and critique their own outputs, often through prompting techniques that encourage introspection and identification of reasoning flaws~\cite{qu2025from, lee-etal-2024-volcano, shinn2023reflexion}, which can be further facilitated by estimating the agent’s own confidence or uncertainty~\cite{bhatt2025know, hu2024uncertainty, chen2025learning}. \textbf{Self-consistency} leverages the generation of multiple candidate outputs, such as diverse reasoning paths or answers, and aggregates them using majority voting or confidence-weighted schemes to select the most reliable results~\cite{wang2022self, liang2023encouraging, wu2025reasoning}. \textbf{Self-questioning} guides the agent to pose and answer critical verification questions grounded in its own reasoning process, enabling the detection of unsupported assertions~\cite{karov2025attentive, ji-etal-2023-towards, manakul2023selfcheckgpt}. Collectively, these strategies establish a foundation for autonomous error detection and correction, empowering agents to self-regulate hallucination risks.

\definecolor{black}{RGB}{0, 0, 0}
\definecolor{gray}{RGB}{234, 234, 234}
\definecolor{light-green}{RGB}{240, 247, 236}
\definecolor{light-blue}{RGB}{237, 242, 248}
\definecolor{light-gray}{RGB}{247, 247, 247}
\definecolor{light-pink}{RGB}{255, 245, 238}
\definecolor{light-orange}{RGB}{252, 241, 232}
\definecolor{light-yellow}{RGB}{252, 248, 232}
\tikzset{%
    parent/.style = {align=center,text width=2.3cm,rounded corners=3pt, line width=0.2mm, fill=gray,draw=black},
    child/.style = {align=center,text width=2.3cm,rounded corners=3pt, fill=blue!10,draw=blue!80,line width=0.3mm},
    grandchild/.style = {align=center,text width=2cm,rounded corners=3pt},
    greatgrandchild/.style = {align=center,text width=1.5cm,rounded corners=3pt},
    greatgrandchild2/.style = {align=center,text width=1.5cm,rounded corners=3pt},    
    referenceblock/.style = {align=center,text width=1.5cm,rounded corners=2pt}, 
    reasoning_top/.style = {align=center,text width=2.0cm,rounded corners=3pt, fill=light-green,draw=black,line width=0.2mm},
    execution_top/.style = {align=center,text width=2.0cm,rounded corners=3pt, fill=light-blue,draw=black,line width=0.2mm},
    perception_top/.style = {align=center,text width=2.0cm,rounded corners=3pt, fill=light-gray,draw=black,line width=0.2mm},
    memorization_top/.style = {align=center,text width=2.0cm,rounded corners=3pt, fill=light-yellow,draw=black,line width=0.2mm},
    communication_top/.style = {align=center,text width=2.0cm,rounded corners=3pt, fill=light-orange,draw=black,line width=0.2mm}, 
    reasoning/.style = {align=center,text width=2.9cm,rounded corners=3pt, fill=light-green,draw=black,line width=0.2mm},
    execution/.style = {align=center,text width=2.9cm,rounded corners=3pt, fill=light-blue,draw=black,line width=0.2mm},
    memorization/.style = {align=center,text width=2.9cm,rounded corners=3pt, fill=light-yellow,draw=black,line width=0.2mm},
    reasoning_work/.style = {align=center, text width=5.2cm, rounded corners=3pt, fill=light-green,draw=black,line width=0.2mm},    
    execution_work/.style = {align=center,text width=5.2cm,rounded corners=3pt, fill=light-blue,draw=black,line width=0.2mm},     
    perception_work/.style = {align=center,text width=8.565cm,rounded corners=3pt, fill= light-gray,draw= black,line width=0.2mm},
    memorization_work/.style = {align=center,text width=5.2cm,rounded corners=3pt, fill= light-yellow,draw= black,line width=0.2mm},
    communication_work/.style = {align=center,text width=8.565cm,rounded corners=3pt, fill= light-orange,draw= black,line width=0.2mm}   
}

\noindent\emph{b) Validator Assistance (VA).} This approach leverages external validators to verify the correctness of an agent’s outputs, aiming to mitigate hallucinations~\cite{cemri2025multi, zhang2025agent}. According to the types of external validators, we divide existing methods into the following five categories: \textbf{Language-based Validators} independently assess the truthfulness or coherence of an agent’s outputs using techniques such as atomic fact decomposition and entailment checking~\cite{jalaian2025hydra, wan-etal-2024-knowledge, xue2025verify}. \textbf{Retrieval-based Validators} rely on some external sources such as search engines to verify whether outputs aligns with established facts~\cite{belyi2024luna}. \textbf{Execution-based Validators} evaluate outputs by running generated codes or plans in external execution environments, enabling direct assessment of correctness through functional outcomes~\cite{ni2023lever}. \textbf{Simulation-based Validators} validate agent behaviors through interaction with sandboxed environments, allowing for realistic testing in tasks involving embodiment, planning, or sequential control~\cite{kleiman2025simulation}. Building on these categories, an increasing number of studies have explored \textbf{Ensemble-based Validators} that integrate multiple types of validators to improve robustness. By enabling cross-verification among different approaches, these methods help mitigate the limitations inherent in individual validation strategies~\cite{wu2025reasoning}. In the end, Table~\ref{table:mitigation} summarizes which types of agent hallucinations can be addressed by the ten listed mitigation methods.

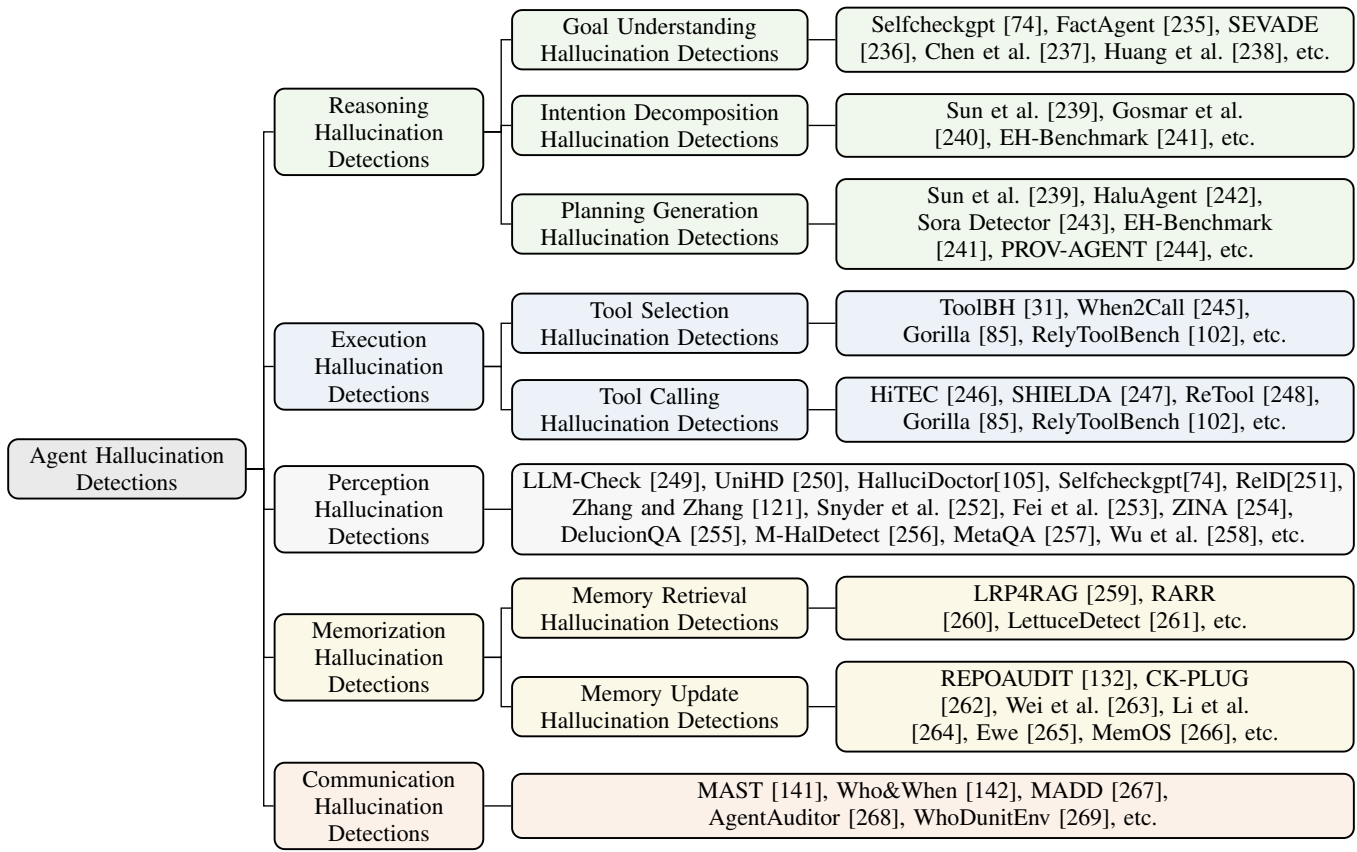
\begin{figure*}[!t]
\centering
{\scriptsize 
\resizebox{\textwidth}{!}{
    \begin{forest}
    for tree={
        forked edges,
        grow'=0,
        draw,
        rounded corners,
        node options={align=center,},
        text width=7cm,
        s sep=6pt,
        calign=edge midpoint,
    }
    [Agent Hallucination \\Detections, fill=gray!45, parent
        [Reasoning \\Hallucination \\Detections, for tree={reasoning_top}
            [Goal Understanding \\Hallucination Detections, reasoning
                [
                    Selfcheckgpt \cite{manakul2023selfcheckgpt}\text{,}
                    FactAgent \cite{li2024large}\text{,}
                    SEVADE \cite{liu2025sevade}\text{,}
                    Chen et al. \cite{chen2025re}\text{,}
                    Huang et al. \cite{huang2025unmasking}\text{,}
                    etc.,
                    reasoning_work
                ]
            ]
            [Intention Decomposition \\Hallucination Detections, reasoning
                [
                    Sun et al. \cite{sun2025towards}\text{,}
                    Gosmar et al. \cite{gosmar2025hallucination}\text{,}
                    EH-Benchmark \cite{pan2025eh}\text{,}
                    etc.,
                    reasoning_work
                ]
            ]
            [Planning Generation \\Hallucination Detections, reasoning
                [
                    Sun et al. \cite{sun2025towards}\text{,}
                    HaluAgent \cite{cheng2024small}\text{,}
                    Sora Detector \cite{chu2024sora}\text{,}
                    EH-Benchmark \cite{pan2025eh}\text{,}
                    PROV-AGENT \cite{souza2025prov}\text{,}
                    etc.,
                    reasoning_work
                ]
            ]
        ]
        [Execution \\Hallucination \\Detections, for tree={execution_top}
            [Tool Selection \\Hallucination Detections, execution
                [
                    ToolBH \cite{r55}\text{,}
                    When2Call \cite{ross2025when2call}\text{,}
                    Gorilla \cite{patil2024gorilla}\text{,}
                    RelyToolBench \cite{r54}\text{,}
                    etc.,
                    execution_work
                ]
            ]
            [Tool Calling \\Hallucination Detections, execution
                [
                    HiTEC \cite{cui2025enhancing}\text{,}
                    SHIELDA \cite{zhou2025shielda}\text{,}
                    ReTool \cite{feng2025retool}\text{,}
                    Gorilla \cite{patil2024gorilla}\text{,}
                    RelyToolBench \cite{r54}\text{,}
                    etc.,
                    execution_work
                ]
            ]
        ]
        [Perception \\Hallucination \\Detections, for tree={perception_top}
            [
                LLM-Check \cite{sriramanan2024llm}\text{,}
                UniHD \cite{chen2024unified}\text{,}
                HalluciDoctor\cite{yu2024hallucidoctor}\text{,}
                Selfcheckgpt\cite{manakul2023selfcheckgpt}\text{,}
                RelD\cite{chen2023hallucination}\text{,}
                Zhang and Zhang \cite{zhang2025hallucination}\text{,}
                Snyder et al. \cite{snyder2024early}\text{,}
                Fei et al. \cite{fei2024fine}\text{,}
                ZINA~\cite{wada2025zina}\text{,}
                DelucionQA~\cite{sadat2023delucionqa}\text{,}
                M-HalDetect~\cite{gunjal2024detecting}\text{,}
                MetaQA~\cite{yang2025hallucination}\text{,}
                Wu et al. \cite{wu2025combating}\text{,}
                etc.,
                perception_work
            ]
        ]
        [Memorization \\Hallucination \\Detections, for tree={memorization_top}
            [Memory Retrieval \\Hallucination Detections, memorization
                [
                    LRP4RAG \cite{hu2024lrp4rag}\text{,}
                    RARR \cite{ross2025rarr}\text{,}
                    LettuceDetect   \cite{kovacs2025lettucedetect}\text{,}
                    etc.,
                    memorization_work
                ]
            ]
            [Memory Update \\Hallucination Detections, memorization
                [
                    REPOAUDIT \cite{Guo2025RepoAuditAA}\text{,}
                    CK-PLUG \cite{bi2025parameters}\text{,}
                    Wei et al. \cite{wei2025llms}\text{,}
                    Li et al. \cite{li2023unveiling}\text{,}
                    Ewe \cite{chen2024improving}\text{,}
                    MemOS \cite{li2025memos}\text{,}
                    etc.,
                    memorization_work
                ]
            ]
        ]
        [Communication \\Hallucination \\Detections, for tree={communication_top}
            [
                MAST \cite{cemri2025multi}\text{,}
                Who\&When \cite{zhang2025agent}\text{,}
                MADD \cite{qiao2025dynamic}\text{,}
                AgentAuditor \cite{luo2025agentauditor}\text{,}
                WhoDunitEnv \cite{barbi2025preventing}\text{,}
                etc.,
                communication_work
            ]
        ]
    ]
\end{forest}   
}}
\caption{A typology of methods of agent hallucination detection. We highlight the representative approaches for each type of agent hallucinations.}
\label{fig:typology_detection}
\end{figure*}

\subsection{Agent Hallucination Detection}
\label{section_4_2}
Based on the introduced taxonomy of agent hallucinations, we summarize the existing detection methods in Fig.~\ref{fig:typology_detection}. From it, we can observe that unlike the mitigation strategies discussed above, research on agent hallucination detection remains relatively limited. Among these detection methods, those addressing perception hallucinations are relatively numerous, whereas methods on memorization hallucinations and communication hallucinations are comparatively limited. We believe this is because perception belongs to the shallow layers of the agent, making hallucination detection and error identification relatively straightforward, which explains the larger amount of corresponding work. In contrast, memory and communication are part of the deeper layers of the agent, where the final outputs are coupled with computations from numerous intermediate modules. This makes hallucination detection and localization more challenging, and consequently, there is comparatively less work in these areas.

\section{Future Directions}
\label{section_5}

LLM-based agent hallucinations represent an emerging research frontier that has attracted increasing attention from both academia and industry. As discussed above, a substantial body of work has focused on the mitigation and detection of agent hallucinations. Building upon our established taxonomy and existing literature, this survey introduces several promising directions for future study. 

\noindent\emph{1) Hallucinatory Accumulation Investigation.} 
Most existing studies investigate hallucination instances and their underlying causes within a single agent loop. However, as we emphasize, agent decision-making is inherently a multi-step and sequential process, in which hallucinations can accumulate and amplify over time. In such cases, hallucinations may initially appear as minor issues, but their iterative accumulation can ultimately lead to severe consequences. Compared to single-step scenarios, hallucination accumulation presents a significantly more complex challenge. Addressing it requires a comprehensive analysis of the agent’s entire decision-making process, thereby facilitating early hallucination detection and mitigation. 

\noindent\emph{2) Accurate Hallucinatory Location.} 
As previously mentioned, in contrast to hallucinations in traditional language models that are typically manifested in textual generation errors, agent hallucinations are far more complex, involving full-chain error propagation across multiple interdependent components. While hallucination taxonomy and attribution we presented in Section~\ref{section_3} have mapped these issues to specific modules within the agent, promptly and accurately locate the source of agent hallucinations in the final outputs remains a significant challenge. This difficulty stems from the fact that agent hallucinations may arise at any stage of the decision-making pipeline and often exhibit complex characteristics such as hallucinatory accumulation and inter-module dependency.  In fact, the limited amount of research on agent hallucination detection in Section~\ref{section_4_2} and the difficulty of conducting such works are also related to this issue. Therefore, future study should focus on designing agent systems capable of modeling and tracing their entire execution trajectories~\cite{zhang2025agent}. For example, lightweight checkpoints can be injected at each stage to verify whether hallucinations have occurred.

\noindent\emph{3) Hallucination Mechanistic Interpretability.}
Mechanistic interpretability (MI) seeks to uncover how hidden representations and internal components of neural networks give rise to specific behaviors~\cite{olah2022mechinterp}. Therefore, MI provides a natural pathway for diagnosing and mitigating hallucinations at the root cause. Recent studies demonstrate that MI techniques, such as feature analysis~\cite{ji2025calibrating} and causal pathway tracing~\cite{jin2024cutting} can reveal the internal sources of hallucinations in LLMs and provide insights for mitigation. However, extending MI to LLM-based agents introduces new challenges. Unlike LLMs, where interpretability analysis is typically performed under single-step prediction with controlled prompt–response settings, agent hallucinations emerge through multi-step sequential interactions involving reasoning, tool use, memorization, and MAS communication. This dynamic process substantially complicates the controlled interventions required for mechanistic analysis. Future research should therefore adapt MI techniques to account for these dynamic and interconnected processes, enabling more precise diagnosis and systematic mitigation of agent hallucinations.

\noindent\emph{4) Unified Benchmark Construction.} 
Existing benchmarks for agent hallucinations are often limited to a specific hallucination type. For example, Guan et al. propose HALLUSIONBENCH to evaluate hallucination issues in visual-language reasoning~\cite{Guan_2024_CVPR}; Hu et al. propose MemoryAgentBench to evaluate hallucination issues in agents during memory retrieval and update~\cite{hu2025evaluating}; Zhang et al. introduce ToolBH to evaluate execution hallucinations in tool use across different scenarios from both depth and breadth perspectives~\cite{r55}. Therefore, there is a lack of a unified benchmark of hallucination evaluation that can define diverse hallucination scenarios and adopt various evaluation metrics to comprehensively assess the extent of hallucinations in agents' reasoning, execution, perception, memory, and communication.

\noindent\emph{5) Continual Self-evolution Capacity.} 
The analysis of agent hallucinations is typically conducted under the assumption of fixed user goals and static environments. However, in practice, both user demands and environmental configurations evolve continuously over time. To remain effective, agents must possess continual self-evolution capabilities that allow them to dynamically adapt to shifting goals and changing conditions. Throughout this process, agents can consistently refine their cognition to mitigate agent hallucinations stemming from outdated knowledge or delayed updates. Based on this important need, integrating the lifelong learning paradigm~\cite{biesialska2020continual} with agents to endow them with more effective dynamic adaptation capabilities represents a promising solution.

\noindent\emph{6) Foundation Architecture Upgrade.} Current LLM-based agents primarily rely on the Transformer architecture. However, this architecture faces challenges in handling long-context information and suffers from high computational complexity, which have gradually revealed performance bottlenecks and contributed to the emergence of hallucination issues. Recent works have explored more effective architectural upgrades, such as introducing linear-complexity modules~\cite{dao2022flashattention} as Transformer components, integrating neural-symbolic systems~\cite{cheng2025neural} to enhance model interpretability through symbolic reasoning, and leveraging automated machine learning~\cite{baratchi2024automated} to design and compose optimal model architectures. In addition, agents typically require a predefined workflow to organize the execution of their various components. While this fixed pattern enhances systematization and controllability, it also reduces flexibility, resulting in limited adaptability and poor scalability. Designing a dynamic self-scheduling agentic system~\cite{li2024autoflow} that can autonomously organize task execution and coordinate multi-agent collaboration is a critical direction for future research.

\section{Conclusion}
\label{section_6}
This paper presents a comprehensive survey of hallucination issues in LLM-based agents, with the goal of consolidating past progress, clarifying current challenges, and outlining future opportunities. We begin by distinguishing agent components into internal states and external behaviors, and, from this perspective, propose a taxonomy of hallucination types occurring at different stages. We then provide an in-depth overview of each type and identify seventeen underlying causes. Furthermore, we summarize ten general approaches for hallucination mitigation, together with corresponding detection methods. Finally, we discuss promising research directions to guide future exploration in this rapidly evolving domain. We believe this survey can serve as a valuable resource for researchers and practitioners, inspiring and facilitating further progress in the development of LLM-based agents.

\bibliographystyle{unsrt}
\bibliography{sample_base}

\clearpage
\appendices

\section{Loop of LLM-based Multi-agent System}
\label{multi_agent_loop}
Different from the single-agent setting, each LLM-based agent in the MAS must communicate with other agents to accomplish goals. To model these processes, a communication structure $G_t$ is introduced among $N$ agents. Furthermore, $G_t$ would evolve with time to align with the dynamic adjustments of MAS. Therefore, compared with Section~\ref{single_agent_loop}, the loop of LLM-based MAS includes two additional procedures: \textbf{Broadcasting} and \textbf{Structure Evolution}. 

\begin{itemize}
    \item \textbf{Reasoning}: Each agent $i$ first generates its own plan $p_t^i$ for the next action conditioned on $b_t^i$ and $g$:
    \begin{equation}
    \label{reasoning_multi_agents}
        p_t^i = \Phi^i (b_t^i,g).
    \end{equation}
    \item \textbf{Execution}: Each agent $i$ then translates $p_t^i$ into an executable action $a_t^i$:
    \begin{equation}
    \label{excution_multi_agents}
        a_t^i = E^i (b_t^i, p_t^i).
    \end{equation}

    \item \textbf{Broadcasting}: The agent $i$ broadcasts its message to its neighbors in $G_t$ according to its plan $p_t^i$: 
    \begin{equation} 
        \big\{q_t^{i \rightarrow j}\ \big|\ j \in N(i) \big\} = B (b_t^i, p_t^i, G_t). 
    \end{equation} 
    
    Here $ \big\{q_t^{i \rightarrow j}\ \big|\ j \in N(i) \big\} $ denotes the messages sent by the agent $i$ to its neighbors (i.e., $N(i)$) at the time step $t$.

    \item \textbf{Feedback}: The learning environment would also provide a reward $r_t$ based on $s_t$ and $\{a_t^i\}_{i=1}^N$: 
    \begin{equation}
        r_t = R \left( s_t, \{a_t^i\}_{i=1}^N \right).
    \end{equation}

    \item \textbf{Environment Transition}: The learning environment transitions to $s_{t+1}$ based on $s_{t}$ and $\{a_t^i\}_{i=1}^N$:
    \begin{equation}
            {\rm Pr}\left(s_{t+1} \big|\ s_t, \{a_t^i\}_{i=1}^N\right) = T\left(s_t, \{a_t^i\}_{i=1}^N\right).
    \end{equation}
   
\item \textbf{Perception}: Each agent $i$ perceives $s_{t+1}$ to generate the observation $ o_{t+1}^i$:
    \begin{equation}
    \label{observation_multi_agent}
            o_{t+1}^i = Z^i\big(s_{t+1}, b_t^i, a_t^i, \big \{q_t^{j \rightarrow i}\ \big|\ j \in N(i) \big\}\big),
    \end{equation}
        where $\big\{q_t^{j \rightarrow i}\ \big|\ j \in N(i) \big\}$ denotes the messages received by the agent $i$ from its neighbors at the time step $t$.

\item \textbf{Memorization}: The external memory module of the agent $i$ is updated as follows,

    \begin{equation}
    \label{memory_multi_agents}
        m_{t+1}^i = L_M^i \left( m_t^i, a_t^i,\big\{q_t^{j \rightarrow i}\ \big|\ j \in N(i) \big\}, o_{t+1}^i \right).
    \end{equation}

\item \textbf{Belief Update}: Then the agent $i$ refines its belief state as follows,
            \begin{equation}
                \label{multi_belief}
                    b_{t+1}^i = L_B^i \left( b_t^i, m_{t+1}^i, a_t^i,\big\{q_t^{j \rightarrow i}\ \big|\ j \in N(i) \big\}, r_t^i, o_{t+1}^i, g \right).
            \end{equation}

\item \textbf{Structure Evolution}: Based on $\{b_{t+1}^i\}_{i=1}^N$ and $G_{t}$, the communication structure can be updated as follows, 
    \begin{equation}
        G_{t+1} = U\left(G_t, \{b_{t+1}^i\}_{i=1}^N\right).
    \end{equation}
\end{itemize}

\begin{figure*}[htbp]
    \centering
    \includegraphics[width=1\textwidth, trim=0cm 0cm 0cm 0cm, clip]{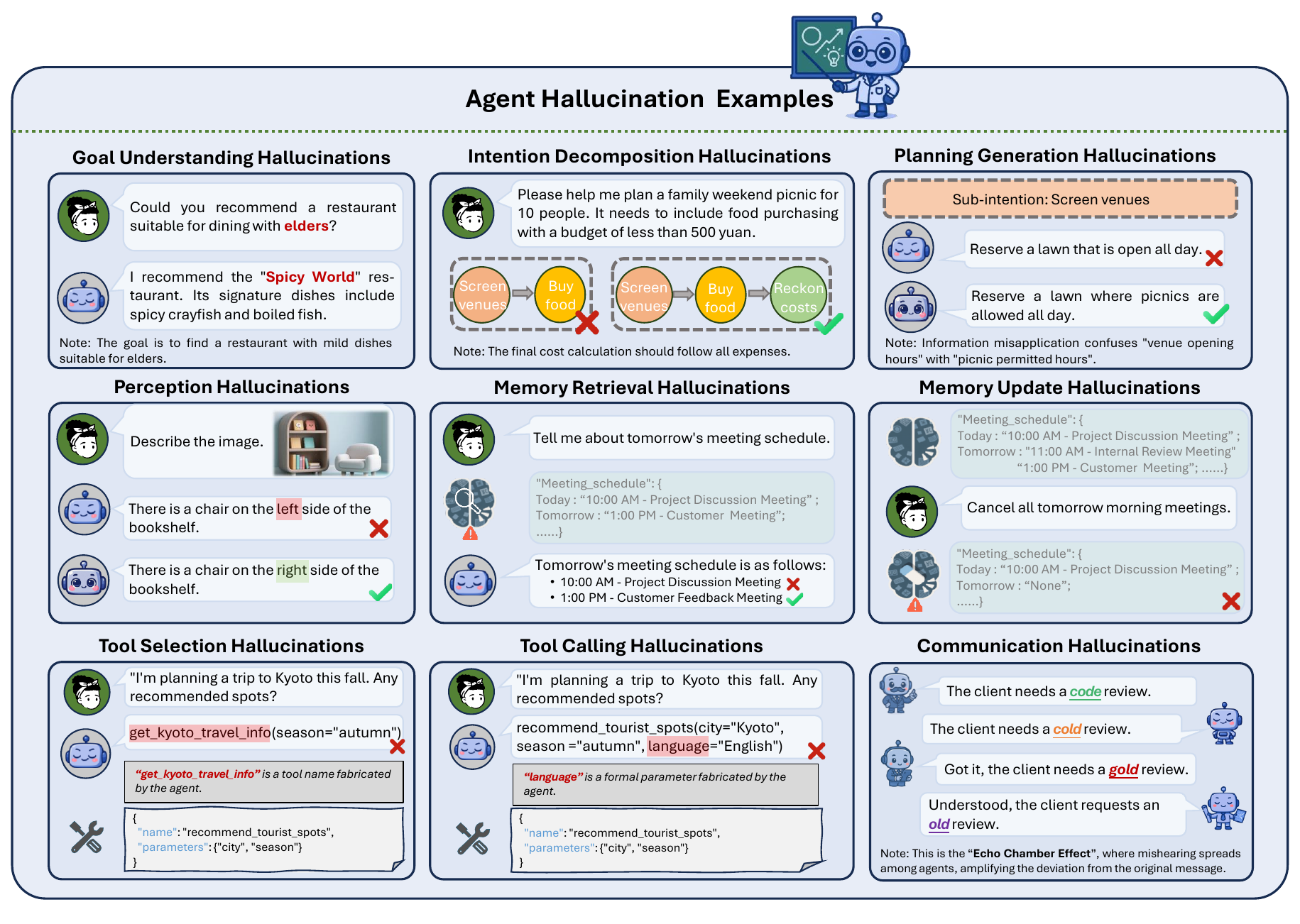} 
    \caption{The depiction of different types of agent hallucinations, each illustrated with a representative example. The detailed explanation is given in Appendix~\ref{Hallucination_Example_Explanation}.}
    \label{fig:Agent Hallucination examples}
\end{figure*}

\section{Hallucination Example Explanation}
\label{Hallucination_Example_Explanation}
As shown in Fig.~\ref{fig:Agent Hallucination examples}, we present the representative examples of each type of agent hallucinations as follows,  

\begin{itemize}
\item \textbf{Goal Understanding Hallucinations.}
In this example, the user asks the agent to ``recommend a restaurant suitable for dining with elders". The underlying goal is clearly to find a restaurant whose cuisine is light and easy to digest and whose environment is appropriate for family gatherings. However, the agent recommends ``Spicy World", a restaurant specializing in very spicy crayfish and boiled fish, thereby completely ignoring the prerequisite ``suitable for elders". 

\item \textbf{Intention Decomposition Hallucinations.} 
In this example, the user asks the agent to plan a ``family picnic for 10 people" and explicitly requires that the total budget for ``purchasing food" must not exceed 500 yuan. Clearly, this plan must include a sub-intentions for ``cost calculation" to verify whether the total expenditure satisfies the budget constraint. However, when decomposing the task, the agent only considers two steps: ``choose a venue" and ``buy food", thereby omitting a crucial component in the decomposition of the user’s intention.

\item \textbf{Planning Generation Hallucinations.}
This example illustrates hallucinations arising in the process of generating a concrete plan for a sub-intention derived from intention decomposition, namely ``screen venues". Given the user’s requirements, the chosen venue should explicitly allow picnicking. Nevertheless, the agent selects merely ``a lawn that is open all day", conflating ``opening hours" with ``time during which picnics are permitted". In other words, the model hallucinates at the planning stage for the sub-intentions, failing to align the generated plan with the constraints specified in the user’s intent.

\item \textbf{Perception Hallucinations.} In this example, the user asks the agent to describe the content of an image. In the image, a chair is located to the right of a bookshelf, yet the agent responds that ``the chair is to the left of the bookshelf". This error is a prototypical perceptual hallucination, in which the agent’s visual understanding or spatial localization is systematically biased.

\item \textbf{Memory Retrieval Hallucinations.}
Here, the user asks the agent to ``tell me about tomorrow’s meeting schedule". The memory store contains two relevant events: a project discussion meeting at 10:00 a.m. today and a customer meeting at 1:00 p.m. tomorrow. However, the agent fails to correctly anchor the events to their respective dates and instead treats both meetings as if they were scheduled for tomorrow. This illustrates that hallucinations can arise at the memory retrieval stage, where agents mis-retrieve or mis-align temporally indexed information.

\item \textbf{Memory Update Hallucinations.}
In this example, the user instructs the agent to ``cancel all meetings tomorrow morning". During the memory update process, however, the agent erroneously clears the entire schedule for the following day, including the afternoon meeting. This reflects a hallucination in self-managed memory editing: the agent performs an overgeneralized or incorrect update, leading to a memory update hallucination.

\item \textbf{Tool Selection Hallucinations.}
The user plans to travel to Kyoto in autumn and asks the agent to recommend tourist attractions. The agent incorrectly calls a non-existent tool named ``get kyoto travel info", whereas the correct tool in the system should be ``recommend tourist spots". This indicates that, when selecting external tools, the agent may ``invent" plausible-sounding but non-existent APIs or function names, thereby exhibiting tool selection hallucinations. 

\item \textbf{Tool Calling Hallucinations.}
Similar to the previous case, this time the agent selects the correct tool, ``recommend tourist spots", but arbitrarily appends an extra parameter language=``English", which is not part of the tool’s actual specification. This demonstrates that even when the tool name is correct, the agent may hallucinate at the parameter level—engaging in hallucinatory argument extension—which can cause the call to fail or yield unintended behavior.

\item \textbf{Communication Hallucinations.}
This example illustrates an ``echo chamber effect" in multi-agent communication, a particularly typical form of communication hallucination. Initially, the client states that they require a ``code review". As the message passes along a chain of agents, the first agent mishears it as ``cold review", the second relays it as ``gold review", and the third finally interprets it as ``old review". Through multiple rounds of transmission, the information is progressively distorted until it becomes completely detached from the original meaning. This hallucination shows that, when multiple agents collaborate, minor mis-hearings or mis-interpretations can be amplified, producing a “telephone game” style accumulation of deviations.  
\end{itemize}

\end{document}